%% file: neurips_2024.tex
\documentclass{article}

\usepackage[final]{neurips_2024}
\input{preamble}

\usepackage[utf8]{inputenc} 
\usepackage[T1]{fontenc}    
\usepackage{hyperref}       
\usepackage{url}            
\usepackage{booktabs}       
\usepackage{amsfonts}       
\usepackage{nicefrac}       
\usepackage{microtype}      
\usepackage{xcolor}         

\usepackage{graphicx}
\usepackage{orcidlink}
\usepackage{sidecap}
\usepackage{wrapfig}
\usepackage{array}
\usepackage{xcolor}
\usepackage{colortbl}
\usepackage{amsmath}
\usepackage{cleveref}
\usepackage{caption}
\usepackage{subcaption}
\usepackage{enumitem}
\usepackage{tabularx}

\usepackage{adjustbox}

\crefname{table}{Table}{Tables}
\crefname{figure}{Figure}{Figures}
\crefname{section}{Section}{Sections}
\crefname{appendix}{Appendix}{Appendices}
\crefname{equation}{Equation}{Equations}
\creflabelformat{equation}{#2#1#3}

\title{Self-Guided Masked Autoencoder}

\author{%
  Jeongwoo Shin$^1$, Inseo Lee$^1$, Junho Lee$^1$, Joonseok Lee$^{1,2}$\thanks{Corresponding author} \\
  $^1$Seoul National University, $^2$Google Research\\
  \texttt{\{swswss, ian.lee, joon2003, joonseok\}@snu.ac.kr} \\
}

\begin{document}

\maketitle

\input{sec/0_abstract}

\input{sec/1_intro}

\input{sec/2_preliminary}
\input{sec/3_motivations}
\input{sec/4_approach}

\input{sec/5_experiments}
\input{sec/6_related_work}
\input{sec/7_conclusion}

\section*{Acknowledgements}

This work was supported by Samsung Electronics Co., Ltd (IO230414-05943-01, RAJ0123ZZ-80SD), by Youlchon Foundation (Nongshim Corp.), and by National Research Foundation (NRF) grants (No.
2021H1D3A2A03038607 / 50\%, RS-2024-00336576 / 10\%, RS-
2023-00222663 / 5\%) and Institute for Information \& communication Technology Planning \& evaluation (IITP) grants (No. RS-2024-00353131 / 25\%, RS-2022-II220264 / 10\%), funded by the government of Korea.

\clearpage
{
    \bibliographystyle{abbrv}
    \bibliography{main}
}
\input{X_suppl}

\end{document}

%% file: preamble.tex
\usepackage[dvipsnames]{xcolor}

\newcommand\rone[1]{\textbf{\textcolor{orange}{LZWD}}}
\newcommand\rtwo[1]{\textbf{\textcolor{blue}{ZFhW}}}
\newcommand\rthree[1]{\textbf{\textcolor{cyan}{BsNN}}}
\newcommand\rfour[1]{\textbf{\textcolor{pink}{Pvc4}}}
\newcommand\rfive[1]{\textbf{\textcolor{green}{3PrK}}}


%% file: sec/0_abstract.tex
\begin{abstract}
Masked Autoencoder (MAE) is a self-supervised approach for representation learning, widely applicable to a variety of downstream tasks in computer vision. In spite of its success, it is still not fully uncovered what and how MAE exactly learns. In this paper, with an in-depth analysis, we discover that MAE intrinsically learns pattern-based patch-level clustering from surprisingly early stages of pre-training. Upon this understanding, we propose \emph{self-guided masked autoencoder}, 
which internally generates informed mask by utilizing its progress in patch clustering, substituting the naive random masking of the vanilla MAE.
Our approach significantly boosts its learning process without relying on any external models or supplementary information, keeping the benefit of self-supervised nature of MAE intact.
Comprehensive experiments on various downstream tasks verify the effectiveness of the proposed method.
\end{abstract}


%% file: sec/1_intro.tex
\section{Introduction}
\label{sec:intro}

Self-supervised learning has been an attractive direction to alleviate the substantial cost for data annotation.
For example, Masked Language Modeling (MLM), predicting masked words of an input sentence, is demonstrated to capture contextual meaning of a word by BERT~\cite{devlin2018bert} and GPT~\cite{brown2020language}.
Motivated from the success of MLM, Masked Image Modeling (MIM) has been introduced in computer vision, utilizing abundant unlabeled image data.
Among them, Masked Autoencoder (MAE)~\cite{he2022masked}, equipped with a Vision Transformer (ViT)~\cite{dosovitskiy2020image}-based asymmetric encoder-decoder structure, demonstrates that simple reconstruction of the RGB pixels for the masked patches is enough to achieve competitive performance on various downstream tasks.

In the wake of MAE's impressive performance, a succession of studies have emerged aiming to augment its capabilities through the integration of informed masking techniques. These innovative endeavors leverage diverse sources of additional information, including attention maps generated by a supervised ViT~\cite{kakogeorgiou2022hide}, knowledge learned by pre-trained self-supervised models~\cite{li2022semmae, chen2023automae}, or supplementary adversarial modules~\cite{shi2022adversarial}, all aiming at refining the quality of the masks. 
However, these prevailing approaches have merely applied informed masking without truly understanding the mechanism of MAE, relying on external resources such as pre-trained models or labels.

To this end, we embark on an in-depth analysis through extensive experiments to understand the internal operation of MAE, as it is still not fully uncovered \emph{what} and \emph{how} MAE exactly learns, despite the several prior endeavors~\cite{cao2022understand, pan2022towards, zhang2022mask, kong2023hierarchical}.
Based on our analysis of MAE, we then explore the potential of MAE to produce informed masks on its own.
We first demonstrate that MAE intrinsically learns \emph{pattern-based patch-level clustering} and this property emerges from \emph{extremely early stages} of pre-training (\cref{sec:analysis:early}).
We then unveil the underlying mechanism of the mask tokens in the decoder (\cref{sec:analysis:decoder}).
Upon this understanding, we propose a novel method to \emph{boost} the training process of MAE via informed masks, generated in an \emph{entirely unsupervised manner} without incurring any external models or supplementary information, unlike the previous informed masking methods.


\cref{fig:intro} illustrates our model compared to the original MAE. Unlike the random masking (b), our method generates informed masks covering the main object entirely (e-f) using the distinguishable patch representations (d)
emerging
from a very early stage of the training.
With the 
internally produced informed masks, MAE accelerates its training process of learning patch-level clustering, leading to clearer and finer embedding space (c, g).
Our contributions are summarized as follows:
\vspace{-0.2cm}
\begin{enumerate}[label=\textbullet, leftmargin=15pt]
    \item We discover that MAE learns \emph{pattern-based patch-level clustering} within each image, emerging from \emph{incredibly early stage} of the pre-training process.

    \item We propose a new masking strategy, \emph{self-guided masked autoencoder}, relying solely on internal quantification of the progress in patch-clustering, free from external models or labels.
    
    
    \item Our comprehensive experiments across various downstream tasks validate that our proposed method genuinely expedites the learning process of MAE.
\end{enumerate}

\input{sec/figure_latex/intro}

%% file: sec/figure_latex/intro.tex
\begin{figure}[t]
    \centering
    \includegraphics[width=1.\linewidth]{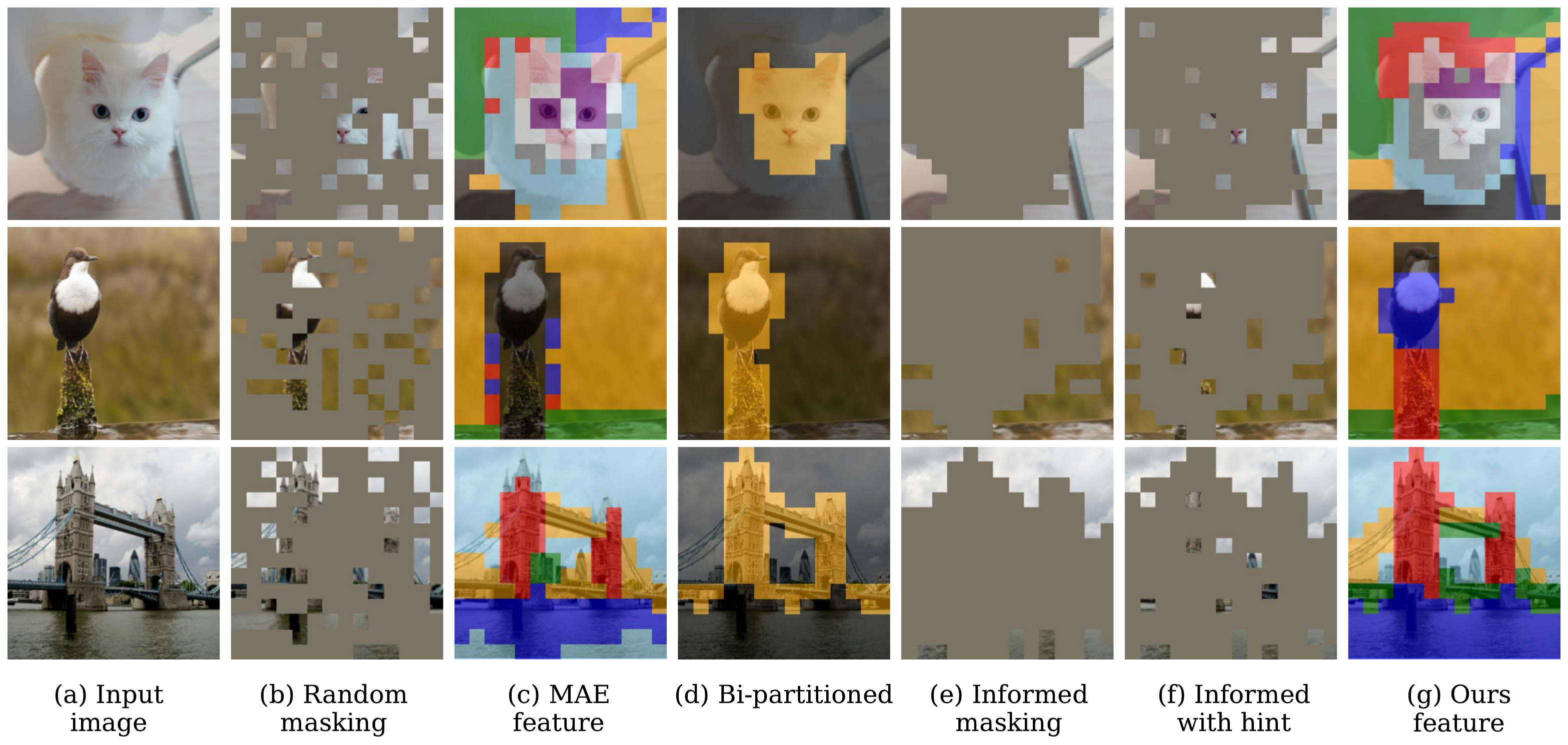}
    \caption{
        \textbf{Illustration of our self-guided MAE}.
    }
    \label{fig:intro}
\end{figure}


%% file: sec/2_preliminary.tex
\section{Preliminary}
\label{sec:preliminary}


\textbf{Masked Autoencoder (MAE).}
MAE~\cite{he2022masked} aims to learn task-agnostic feature representations for various downstream vision tasks, \emph{e.g.}, classification, detection, or segmentation.

Given an image of size $H \times W$, 
MAE first splits it into
to same-sized $P \times P$ image patches.
Each patch is linearly mapped to a $d$-dimensional embedding.
As a result, the input image is represented as a set of these features, denoted by $\mathcal{X} = \{\mathbf{x}^{(1)}, ..., \mathbf{x}^{(n)} : \mathbf{x}^{(i)} \in \mathbb{R}^d \}$, where $n = HW/P^2$ is the number of patches.
We call $\mathbf{x}^{(i)}$ as a `patch' or `token' embedding interchangeably.
MAE randomly masks out a subset of $n$ patches in $\mathcal{X}$.
The set of masked and visible patches are denoted by $\mathcal{X}_m$ and $\mathcal{X}_v$, respectively, where $\mathcal{X}_m \cup \mathcal{X}_v = \mathcal{X}$ and $\mathcal{X}_m \cap \mathcal{X}_v = \phi$.


MAE adopts an asymmetric encoder-decoder structure based on ViT~\cite{dosovitskiy2020image}.
The encoder $E$ takes $\mathcal{X}_v$ as input and produces a same-sized set of embeddings, denoted by $\mathcal{X}'_v = \{\mathbf{x}'^{(i)} : \mathbf{x}'^{(i)} \in E(\mathcal{X}_v)\}$.
Through the encoding, the patch representations are updated to reflect the context of the entire image, only from the visible parts.
Then, the decoder $D$ takes a set of $n$ patch embeddings, denoted by
$\tilde{\mathcal{X}} = \{\tilde{\mathbf{x}}^{(i)} : \tilde{\mathbf{x}}^{(i)} = \mathbf{m} ~ \text{if} ~ \mathbf{x}^{(i)} \in \mathcal{X}_m, \tilde{\mathbf{x}}^{(i)} = \mathbf{x}'^{(i)} ~ \text{if} ~ \mathbf{x}^{(i)} \in \mathcal{X}_v \}$, as input, where $\mathbf{m} \in \mathbb{R}^d$ is a learnable mask token.
Each $\mathbf{x} \in \mathcal{X}_m$ is substituted with a mask token $\mathbf{m}$, and a corresponding positional encoding is applied to distinguish them.
The decoder targets to reconstruct the raw RGB pixels of $\mathcal{X}_m$.
Once trained, only the encoder is deployed for downstream tasks.

\textbf{Hierarchical Latent Variable Model.}
Kong \emph{et al.}~\cite{kong2023hierarchical} recently discovers that the internal operation of MAE can be explained under the framework of a hierarchical latent variable model.
There exists high-level shared information $\mathbf{c}$ in the input image, and it is equivalent to statistical dependency among the patches in $\mathcal{X}$.
The MAE encoder $E(\mathcal{X}_v)$ learns high-level latent variables by estimating the shared information $\hat{\mathbf{c}}$ from the visible patches $\mathcal{X}_v$,
and the decoder $D([E(\mathcal{X}_v); \mathbf{m}])$ performs the reconstruction task by inducing $\mathcal{X}_m$ from $\hat{\mathbf{c}}$ via the mask tokens. 

%% file: sec/3_motivations.tex
\newcommand{\newhref}[2]{\hyperref[#1]{\ref*{#1}#2}}
\newcommand{\newhcref}[2]{\hyperref[#1]{\cref*{#1}#2}}

\section{Analysis of MAE}
\label{sec:analysis}

We study what and how MAE learns with a concept of token relation (\cref{sec:analysis:token_relation}), and demonstrate that it learns pattern-based patch-level clustering (\cref{sec:analysis:MAE_clustering}) from early stages of training (\cref{sec:analysis:early}).
We then illuminate the underlying mechanism of the
MAE decoder (\cref{sec:analysis:decoder}). 
In this section, we use the ViT-B MAE~\cite{he2022masked} pre-trained for 400 epochs on ImageNet-1K~\cite{deng2009imagenet} and all experiments have been conducted on 10\% of ImageNet-1K training set, unless noted otherwise.


\subsection{Token Relation}
\label{sec:analysis:token_relation}

In order to understand what MAE learns, we analyze its token embeddings
with their quantified pair-wise relationships, \emph{i.e.,} \emph{attention score matrix} $\mathbf{A}$ and \emph{cosine similarity matrix} $\mathbf{M}$.
For the input patches $\mathbf{X} \in \mathbb{R}^{n \times d}$ and the transformer weights $\mathbf{W}^{\{Q, K, V\}} \in \mathbb{R}^{d \times d'}$ for queries, keys, and values, respectively, $\mathbf{A} \in \mathbb{R}^{n \times n}$ and $\mathbf{M} \in \mathbb{R}^{n \times n}$ are given by

\vspace{-.5cm}
\begin{tabularx}{\textwidth}{@{}XX@{}}
  \begin{equation}
    \mathbf{A} = \text{Softmax}(\mathbf{XW}^Q (\mathbf{XW}^K)^\top / \sqrt{d'}),
    \label{eq:attention_score}
  \end{equation}
\vspace{-0.25cm} &
  \begin{equation}
  \mathbf{M}_{i,j} = \frac{{\mathbf{x}'_i}^\top \mathbf{x}'_j}{\| \mathbf{x}'_i \|_2 \cdot \| \mathbf{x}'_j \|_2},
    \label{eq:cosine_similarity}
  \end{equation}
\vspace{-1.2cm}
\end{tabularx}
where $\mathbf{X}' = \mathbf{A}(\mathbf{XW}^V) \in \mathbb{R}^{n \times d'}$ and $\mathbf{x}'_i \in \mathbb{R}^{d'}$ indicates the $i$-th row of $\mathbf{X}'$.

We present analysis with $\mathbf{A}$ and $\mathbf{M}$ under two settings.
First, we calculate them using complete set of patches $\mathcal{X}$ at the encoder as $E(\mathcal{X})$.
This ideal setting offers the most accurate $\hat{\mathbf{c}}$, suitable to analyze the features learned by MAE.
As an alternative, they can be obtained from the practical setting of the decoder $D([E(\mathcal{X}_v); \mathbf{m}])$ which contains mask tokens.
Since only the visible tokens are exploited to estimate $\hat{\mathbf{c}}$, this setting would produce less accurate token relations compared to the former one.

\subsection{What is Learned by MAE?}
\label{sec:analysis:MAE_clustering}





We investigate the 
distribution of patch relationships in the learned embedding space,
using
the last layer embeddings of $E(\mathcal{X})$ and $D([E(\mathcal{X}_v); \mathbf{m}])$ for 
196 $(14 \times 14)$ 
patches of set-aside test images.

\input{sec/figure_latex/mae_moco_vit}

\textbf{Qualitative Analysis.}
In \cref{fig:mae_moco_vit}, we compare the patch representations among different models (MAE, MoCo~\cite{he2020moco}, and ViT~\cite{dosovitskiy2020image}).
Figure \newhref{fig:mae_moco_vit}{a} depicts the normalized pairwise cosine similarity matrix ($\mathbf{M}$) for all $196 \times 196$ patch pairs for a test image.
The MAE encoder shows more polarized values, \emph{i.e.}, higher variance, indicating that patches are more clearly clustered.
Figure \newhref{fig:mae_moco_vit}{b} illustrates the cosine similarity between the mean of patches and all individual patches. 
In the examples in \cref{fig:mae_moco_vit}, the background patches are majority, so the mean patch is closer to the background. Patches corresponding to the main object clearly show lower similarity to the mean (background), indicating that the MAE encoder has learned patch clustering based on \emph{visual patterns}, \emph{i.e.}, texture and color.
Similar results in the projected latent space are provided in \cref{appendix:convergence}.
Figure \newhref{fig:mae_moco_vit}{c} shows the attention scores of the class (\texttt{[CLS]}) token.
As the class token is not updated during self-supervised training, it does not carry particularly meaningful information and therefore could be regarded as a random vector. As a result,
the class token does not lean towards any specific patch under self-supervision, and thus the score is distributed similarly to the relationship with the mean patch in Figure \newhref{fig:mae_moco_vit}{b}.
In contrast, MoCo and ViT fail to clearly distinguish the patterns among the whole patches. 

Despite the limited information $\mathcal{X}_v$, the decoder also exhibits proficiency in grouping patches based on their patterns, albeit not as effective as the encoder.

\textbf{Quantitative Analysis.}
We additionally measure the feature variance ($\sigma_F$) and variance of the pairwise similarities ($\sigma_S$), on the ImageNet-1K validation set:
\begin{equation}
  \sigma_F = \frac{1}{n} \sum_{i=1}^n (\frac{\mathbf{x}_i}{\|\mathbf{x}_i\|_2} - \bar{\mathbf{x}})^2, \
  \sigma_S = \sum_{i=1}^n \sum_{j \ne i} \frac{(\mathbf{M}_{i,j} - \bar{\mathbf{M}})^2}{n(n-1)},
  \label{eq:feature_sim_variance}
\end{equation} 
where $\bar{\mathbf{x}} = \frac{1}{n} \sum_i \mathbf{x}_i / \|\mathbf{x}_i\|_2$ and $\bar{\mathbf{M}} = \frac{1}{n(n-1)} \sum_{i,j} \mathbf{M}_{i,j}$.
Higher $\sigma_F$ indicates patch embeddings are spread out more widely in the feature space, while higher $\sigma_S$ indicates stronger patch clustering.

\input{sec/table_latex/mae_moco_vit}
In \cref{tab:mae_moco_vit},
the MAE encoder and decoder show significantly higher $\sigma_F$ and $\sigma_S$ compared to MoCo and ViT, suggesting that their patch embeddings are more diversely clustered in the embedding space rather than in a simpler alternative, \emph{e.g.}, bi-partition.
Given the significant utilization of high-frequency information (\emph{e.g.}, pattern or texture) in MAE (\cref{fig:analysis}), we can quantitatively confirm that MAE effectively clusters patches based on their patterns.
MoCo and ViT show significantly lower $\sigma_F$ and $\sigma_S$, as they tend to learn a simpler form of feature maps, 
aligned with~\cite{park2023selfvit}.
To alleviate the concern that the large variance might be a result of a few extremely clustered features, instead of good separability, we additionally measure Normalized Mutual Information (NMI)~\cite{strehl2002nmi} between queries and keys, which is an indicator of homogeneity in attention map~\cite{park2023selfvit}.
As shown in \cref{fig:attn_nmi}, we confirm with non-zero NMI that MAE does not collapse to a few extremely separated feature groups.


\textbf{Summary.}
MAE learns patch-level clustering in an image based on their visual patterns. Operating only with visible tokens, the decoder learns a similar but less clearer trend than the encoder.



\subsection{When Does MAE Learn Patch Clustering?}
\label{sec:analysis:early}

Given that the MAE learns patch clustering upon completion of pre-training, when does it start to learn them in pre-training?
We answer this question by tracking the token relations of MAE.


\textbf{Evolving Bi-partitioning across Training.}
We start with the simplest form of token clusters, \emph{i.e.}, bi-partitioning.
We cluster the patches into the two most prominent sub-groups 
by applying graph-cut to $\mathbf{M}$ from the final layer.
Based on this clustering, we trace the mean of inter-cluster edge weights ($\mu_\text{inter}$) and mean of intra-cluster edge weights ($\mu_\text{intra}$) with $\mathbf{M}$ and $\mathbf{A}$, across the training.


Figure \newhref{fig:bipartition_kld}{a} shows $\mu_\text{inter}$ and $\mu_\text{intra}$ measured with $\mathbf{M}$ and $\mathbf{A}$.
We observe two notable patterns regarding the gap $\mu_\text{intra} - \mu_\text{inter}$:
1) \emph{the gap tends to get larger} along with the training steps, more prominently with the attention scores.
2) there is a clear margin between $\mu_\text{intra}$ and $\mu_\text{inter}$ \emph{from very early stages}.
The decoder also shows a similar but less prominent trend. 


\input{sec/figure_latex/bipartition_kld}

\textbf{Convergence of Token Relations.}
Going beyond investigating token clusters, we directly track the gap between the distribution of token relations during the training \emph{vs.} upon completion. 
Specifically, we consider the mean KL divergence $\delta_{i}(j)$ in the $i$-th layer at the $j$-th epoch over a set of images $\mathcal{D}$:
\begin{equation}
  \delta_{i,j} = \frac{1}{|\mathcal{D}|} \sum_{\mathbf{I} \in \mathcal{D}} D_\text{KL}(\mathbf{R}^{(i)}_{j}(\mathbf{I}) \| \mathbf{R}^{(i)}_{N}(\mathbf{I})),
  \label{eq:kld}
\end{equation}
where $N$ is total epochs and $\mathbf{R}^{(i)}_{j}: \mathbb{R}^{H \times W \times 3} \rightarrow \mathbb{R}^{n\times n}$ is a function mapping an input image $\mathbf{I}$ to a token relation matrix (\emph{e.g.}, $\mathbf{M}$ or $\mathbf{A}$) computed with the $i$-th layer embeddings at the $j$-th epoch.

Figure \newhref{fig:bipartition_kld}{b} depicts $\delta_{i}(j)$ for even-numbered layers up to 400 epochs,
measured with $\mathbf{M}$ and $\mathbf{A}$.
It clearly shows that $\delta_{i}$ monotonically decreases,
converging quickly at early epochs, indicating that the patches begin to be clustered from early epochs.
This result strongly implies that MAE learns the token relations at early epochs and progressively strengthens it along the rest of training.
The decoder also shows a similar trend, but with less prominence.

\textbf{Summary.} MAE learns
to cluster the patches from the early stage of training.


\subsection{Operations of the Decoder}
\label{sec:analysis:decoder}

In previous experiments, we observe that the decoder $D([E(\mathcal{X}_v); \mathbf{m}])$ in the practical setting is still able to build complete token relation, which verifies that the decoder exploits the estimated shared information $\hat{\mathbf{c}}$ conveyed from the encoder $E(\mathcal{X}_v)$ to complement the missing information in masked-out tokens ($\mathcal{X}_m$) and reconstruct them.
Connecting this to our discovery in \cref{sec:analysis:MAE_clustering}, we claim that the pattern-based patch clustering learned by MAE conceptually corresponds to this $\hat{\mathbf{c}}$.
If the encoder is trained sufficiently, its output embeddings $\mathcal{X}'_v$ for the visible tokens would convey the general context (\emph{i.e.}, $\hat{\mathbf{c}}$) of the entire image.
Then, through the decoding process, mask tokens are contextualized by selectively attending to $\mathcal{X}'_v$, thereby possessing the essential information to represent the target patches $\mathcal{X}_m$, originally derived from $\hat{\mathbf{c}}$.
Therefore, by reversing this process, we can assess if the encoder has been sufficiently trained to precisely associate the patches by quantifying $\hat{\mathbf{c}}$ deployed in $D([E(\mathcal{X}_v); \mathbf{m}])$, which is estimated by $E(\mathcal{X}_v)$.
Based on this idea, we propose a novel metric to measure it during training, which will be the key to our proposed method in \cref{sec:approach}.



\textbf{Exploitation Rate.}
We pose that the overall attention weight on mask tokens at the decoder is a good indicator to quantify the amount of $\hat{\mathbf{c}}$ utilized by the decoder.
Specifically, we define the \emph{exploitation rate} of the mask tokens over the decoder layers using the attention score matrix 
$\mathbf{A}$ (\cref{eq:attention_score}),
which can be interpreted as the special case of attention rollout~\cite{abnar2020attnrollout}. 
For the sets of token indices $\mathcal{A}$ and $\mathcal{B}$, the \textit{exploitation rate} $r^{(l)}_{\mathcal{A} \rightarrow \mathcal{B}} \in \mathbb{R}$ of the tokens in $\mathcal{A}$ to construct the tokens in $\mathcal{B}$ at the $l$-th layer is defined as the average attention weights relying on the tokens in $\mathcal{A}$ for the tokens in $\mathcal{B}$:
\begin{equation}
  r^{(l)}_{\mathcal{A} \rightarrow \mathcal{B}} = \frac{1}{|\mathcal{B}|}\sum_{i \in \mathcal{B}}\sum_{j \in \mathcal{A}} \mathbf{A}^{(l)}_{ij},
\end{equation}
where $l = 1, ..., L$ is the layer index, and $\mathbf{A}^{(l)}$ is the attention score matrix 
at the $l$-th layer.
For $\mathcal{A}$ and $\mathcal{B}$, we are interested in the set of mask tokens $\mathcal{M} = \{i : \mathbf{x}^{(i)} \in \mathcal{X}_m \}$, of visible tokens $\mathcal{V} = \{i : \mathbf{x}^{(i)} \in \mathcal{X}_v \}$, and of all tokens $\mathcal{O} = \{1, ..., n\}$.
For example, $r^{(l)}_{\mathcal{M} \rightarrow \mathcal{V}} = 0.7$ indicates that the contextualized visible tokens consist of mask tokens (70\%) and visible tokens (30\%) on average.

Then, we recursively accumulate these ratio across all layers to get the overall exploitation rate of the mask tokens.
Formally, the \emph{accumulated exploitation rate} $R^{(l)}_{\mathcal{A} \rightarrow \mathcal{B}} \in \mathbb{R}$ of the tokens in a set $\mathcal{A}$ to construct the tokens in a set $\mathcal{B}$ \emph{up to} the $l$-th layer is defined by
\begin{equation}
  R^{(l)}_{\mathcal{A} \rightarrow \mathcal{B}} = r^{(l)}_{\mathcal{A} \rightarrow \mathcal{B}} \cdot R^{(l-1)}_{\mathcal{A} \rightarrow \mathcal{A}} + r^{(l)}_{\mathcal{B} \rightarrow \mathcal{B}} \cdot R^{(l-1)}_{\mathcal{A} \rightarrow \mathcal{B}},
  \label{formula_six}
\end{equation}
where $R^{(0)}_{\mathcal{A} \rightarrow \mathcal{B}} = 1$. 
At the $l$-th layer, tokens in $\mathcal{B}$ (denoted by $\mathcal{B}^{l}$) consist of tokens in both $\mathcal{A}$ and $\mathcal{B}$ from the previous layer, $\mathcal{A}^{l-1}$ and $\mathcal{B}^{l-1}$, with their respective ratios of {$r^{(l)}_{\mathcal{A} \rightarrow \mathcal{B}}$} and {$r^{(l)}_{\mathcal{B} \rightarrow \mathcal{B}}$}.
Thus, the left term {$r^{(l)}_{\mathcal{A} \rightarrow \mathcal{B}} \cdot R^{(l-1)}_{\mathcal{A} \rightarrow \mathcal{A}}$} means the \emph{overall exploitation rate} of tokens in $\mathcal{A}$ to construct those in $\mathcal{B}^{l}$, coming through $\mathcal{A}^{l-1}$, and similarly, {$r^{(l)}_{\mathcal{B} \rightarrow \mathcal{B}} \cdot R^{(l-1)}_{\mathcal{A} \rightarrow \mathcal{B}}$} indicates that coming through $\mathcal{B}^{l-1}$.

Finally, the proportion of information from the visible tokens ($R^{(l)}_{\mathcal{V} \rightarrow \mathcal{O}}$) and that from the mask tokens ($R^{(l)}_{\mathcal{M} \rightarrow \mathcal{O}}$) after $l$-th layer with masking ratio of $m$ are given by
\begin{equation}
  R^{(l)}_{\mathcal{V} \rightarrow \mathcal{O}} = m \cdot R^{(l)}_{\mathcal{V} \rightarrow \mathcal{M}} + (1-m) \cdot R^{(l)}_{\mathcal{V} \rightarrow \mathcal{V}}, \quad
  R^{(l)}_{\mathcal{M} \rightarrow \mathcal{O}} = m \cdot R^{(l)}_{\mathcal{M} \rightarrow \mathcal{M}} + (1-m) \cdot R^{(l)}_{\mathcal{M} \rightarrow \mathcal{V}}.
  \label{eq:accum_exploit}
\end{equation}

\input{sec/figure_latex/mask_ratio_new}
\textbf{Empirical Analysis.}
We measure the exploitation rate of visible tokens ($R^{(l)}_{\mathcal{V} \rightarrow \mathcal{O}}$) and mask tokens ($R^{(l)}_{\mathcal{M} \rightarrow \mathcal{O}}$) in \cref{fig:mask_ratio_new}.
Surprisingly, $R^{(l)}_{\mathcal{M} \rightarrow \mathcal{O}}$ surpasses $R^{(l)}_{\mathcal{V} \rightarrow \mathcal{O}}$ after some moment, denoted by $T$.
Heavy exploitation on the mask tokens in every decoder layer strongly indicates that they truly hold substantial amount of shared information estimated by the encoder, which is more valuable than simple interpolation of visible patches (before $T$ epochs) to represent masked out patches.
We observe $T \approx 50$, but this may differ depending on the model or dataset.

\textbf{Summary.}
When the encoder is sufficiently trained to cluster patches, the encoder outputs reflect the shared information, and they are utilized to constitute mask tokens in the decoder. This means the mask tokens possess this patch clustering information and start to be intensely exploited for reconstructing masked-out patches. 
Thus, we can conversely infer from a high exploitation rate of the mask tokens in the decoder that the mask tokens have patch clustering information conveyed from the encoder, sufficiently to cluster the patches.
This process is verified via measuring the shared information learned by the encoder $E(\mathcal{X}_v)$ during training by tracking the accumulated exploitation rate in \cref{eq:accum_exploit}. Heavy exploitation of the mask tokens in decoder after $T \approx 50$ implies that the encoder is presently trained sufficiently to cluster the patches.



%% file: sec/figure_latex/mae_moco_vit.tex
\begin{figure*}[t]
    \begin{minipage}{0.1\textwidth}%
        \begin{subfigure}{\textwidth}%
            \includegraphics[width=\linewidth]{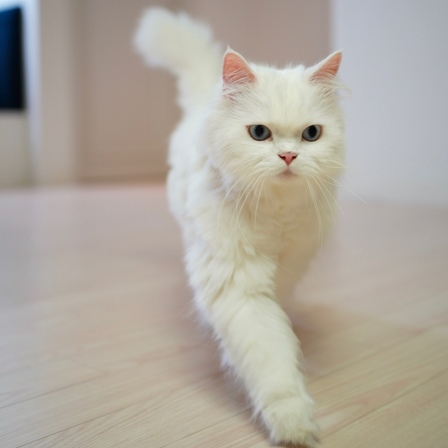}
            \centering
            \scriptsize{\color{white}{...}\color{black}Image 1}
            \newline
            \newline
        \end{subfigure}
        \begin{subfigure}{\textwidth}%
            \includegraphics[width=\linewidth]{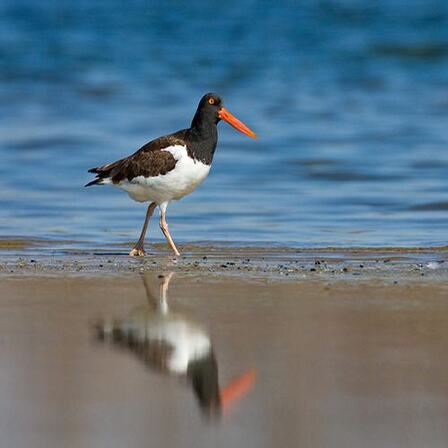}
            \centering
            \scriptsize{\color{white}{.}\color{black}Image 2}
        \end{subfigure}
    \end{minipage}
    \hspace{0.5mm}
    \begin{minipage}{0.44\textwidth}%
        \begin{subfigure}{\linewidth}%
            \includegraphics[width=\textwidth]{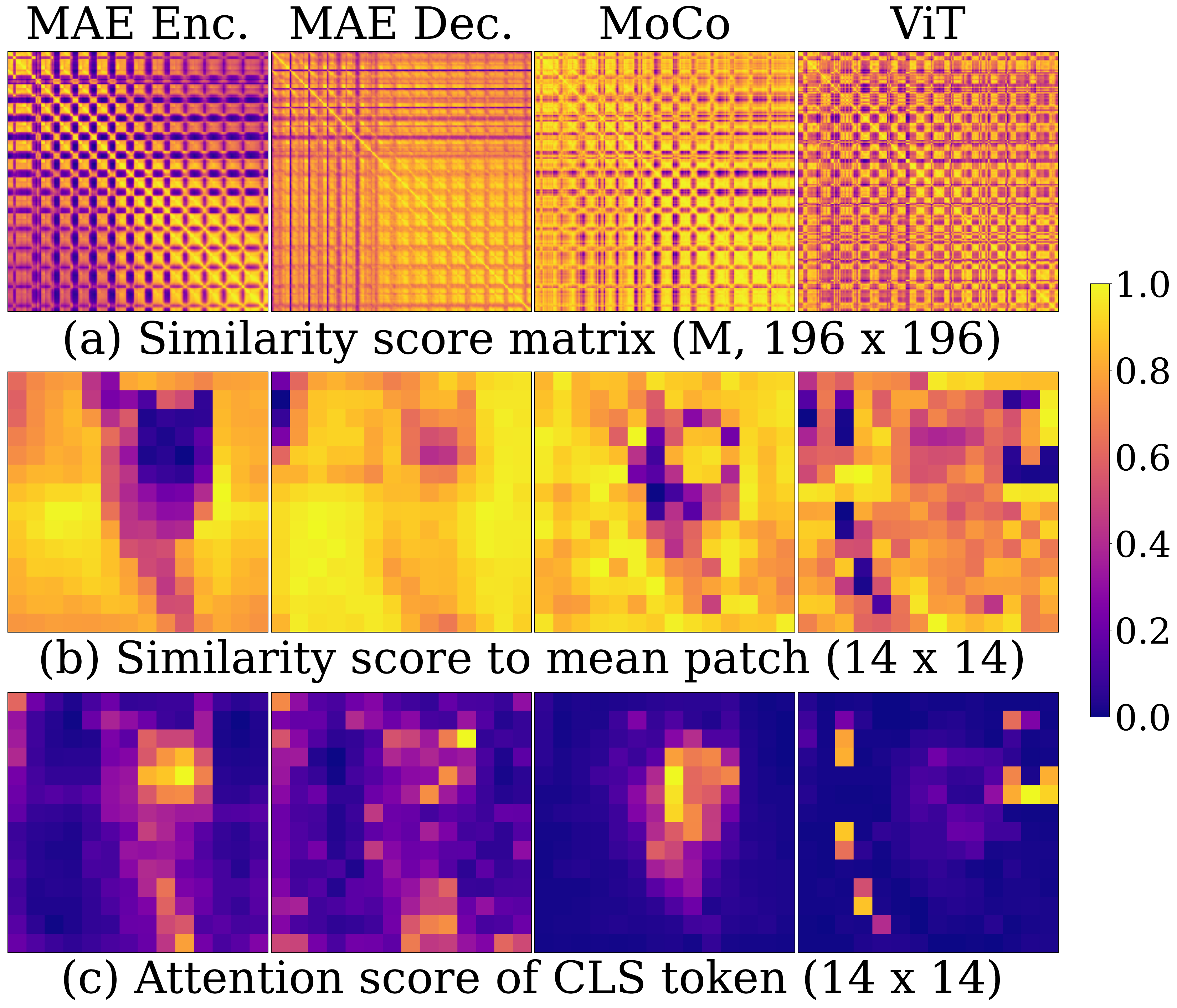}
            \caption*{Representations of image 1}
        \end{subfigure}
    \end{minipage}%
    \begin{minipage}{0.44\textwidth}%
        \begin{subfigure}{\linewidth}%
            \includegraphics[width=\textwidth]{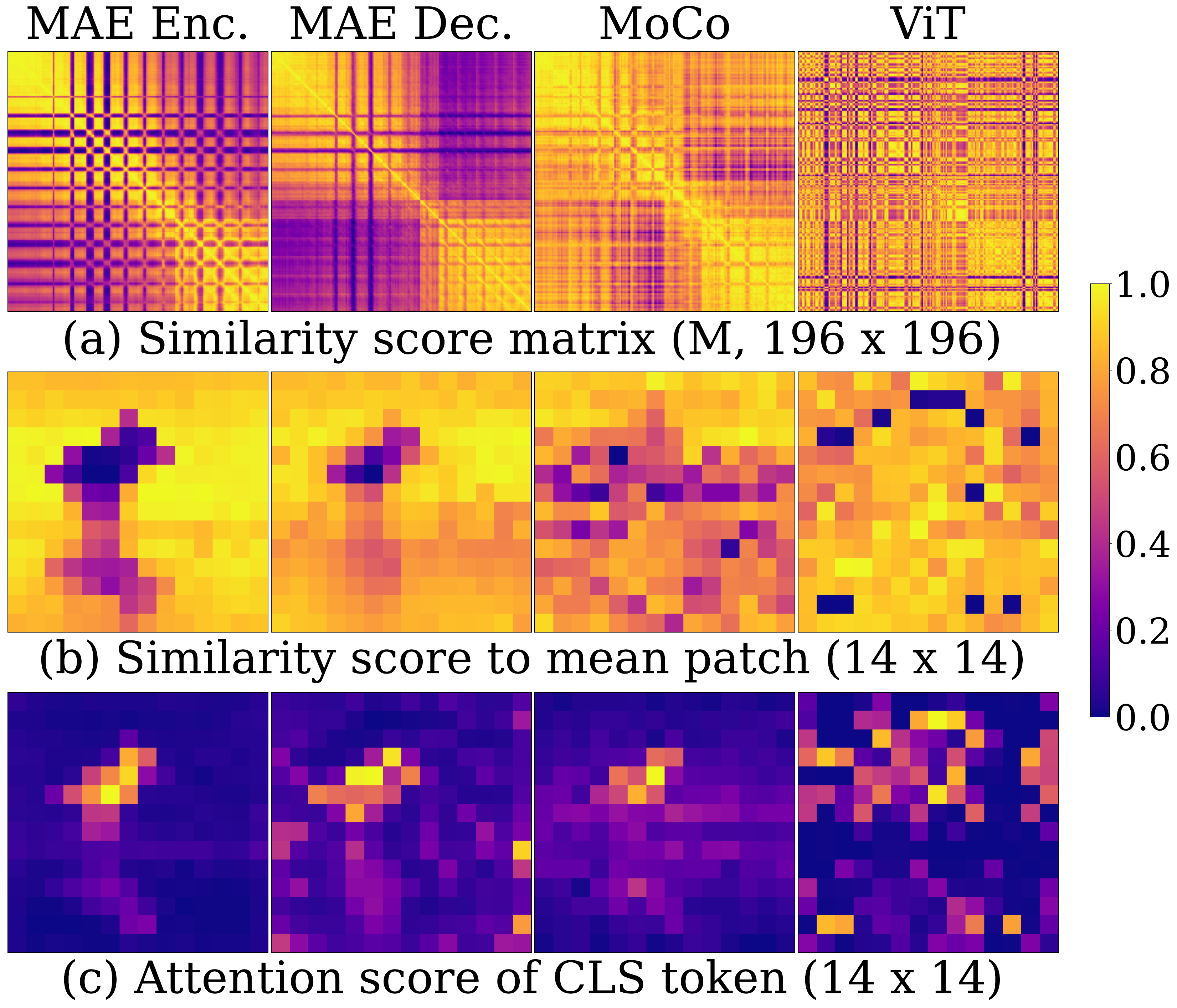}
            \caption*{Representations of image 2}
        \end{subfigure}
    \end{minipage}%
    \caption{\textbf{Relationships among the patch embeddings.} (a) Pairwise similarity matrix for all $196 \times 196$ pairs of patches.
    (b) Similarity between the mean patch and  all individual patches. (c) Attention score of the class token.}
    \label{fig:mae_moco_vit}
\end{figure*}

%% file: sec/table_latex/mae_moco_vit.tex

\newcolumntype{C}[1]{>{\centering\arraybackslash}p{#1}}
\begin{wraptable}{r}{.41\linewidth}
\vspace{-0.43cm}
\centering
    \captionsetup{width=.95\linewidth}
    \caption{Feature variance ($\sigma_F$) and similarity variance ($\sigma_S$).}
    \vspace{-0.1cm}
    \footnotesize
    \begin{tabular}{l|C{1.25cm}|C{1.25cm}} 
        \toprule
        Feature & $\mathbb{E}[\sigma_F]$ & {$\mathbb{E}[\sigma_S]$} \\ 
        \midrule
        MAE encoder~~ & 0.08 & 0.075 \\
        MAE decoder & 0.11 & 0.059 \\
        MoCo~\cite{he2020moco} & 0.01 & 0.003 \\ 
        ViT~\cite{dosovitskiy2020image} & 0.02 & 0.012 \\
        \bottomrule
    \end{tabular}
    \label{tab:mae_moco_vit}
\vspace{-.4cm}
\end{wraptable}


%% file: sec/figure_latex/bipartition_kld.tex
\begin{figure}[t]
    \centering
        \includegraphics[width=0.75\linewidth]{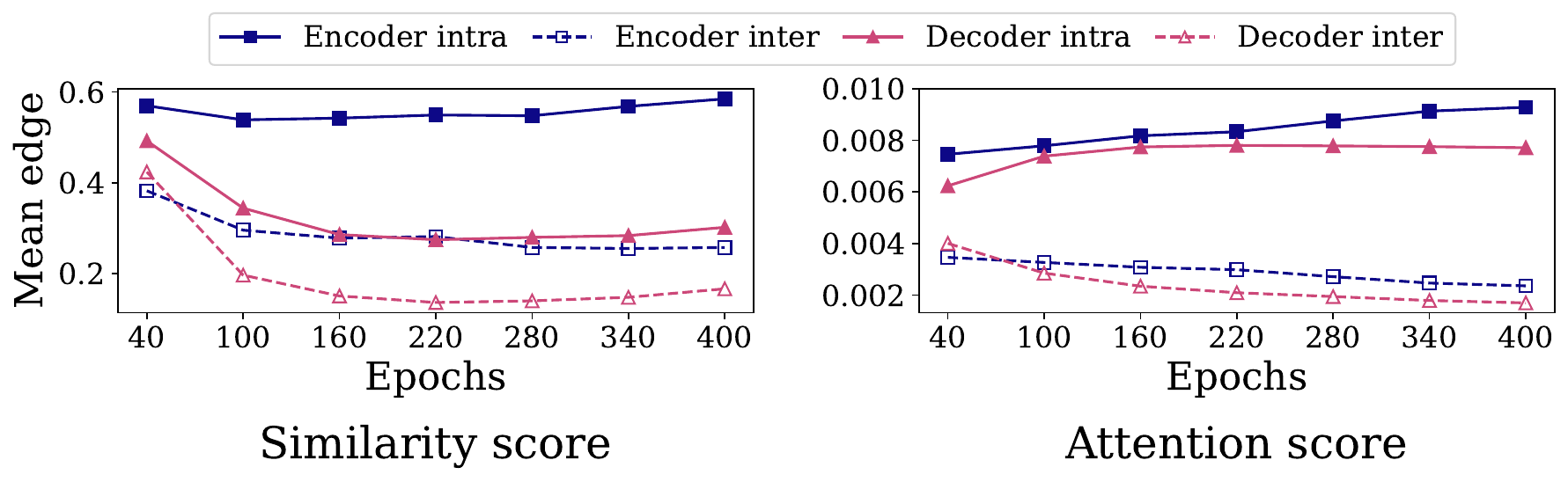}
        \caption*{(a) Bi-partitioning performance}
        \vspace{2pt}
        \includegraphics[width=0.75\linewidth]{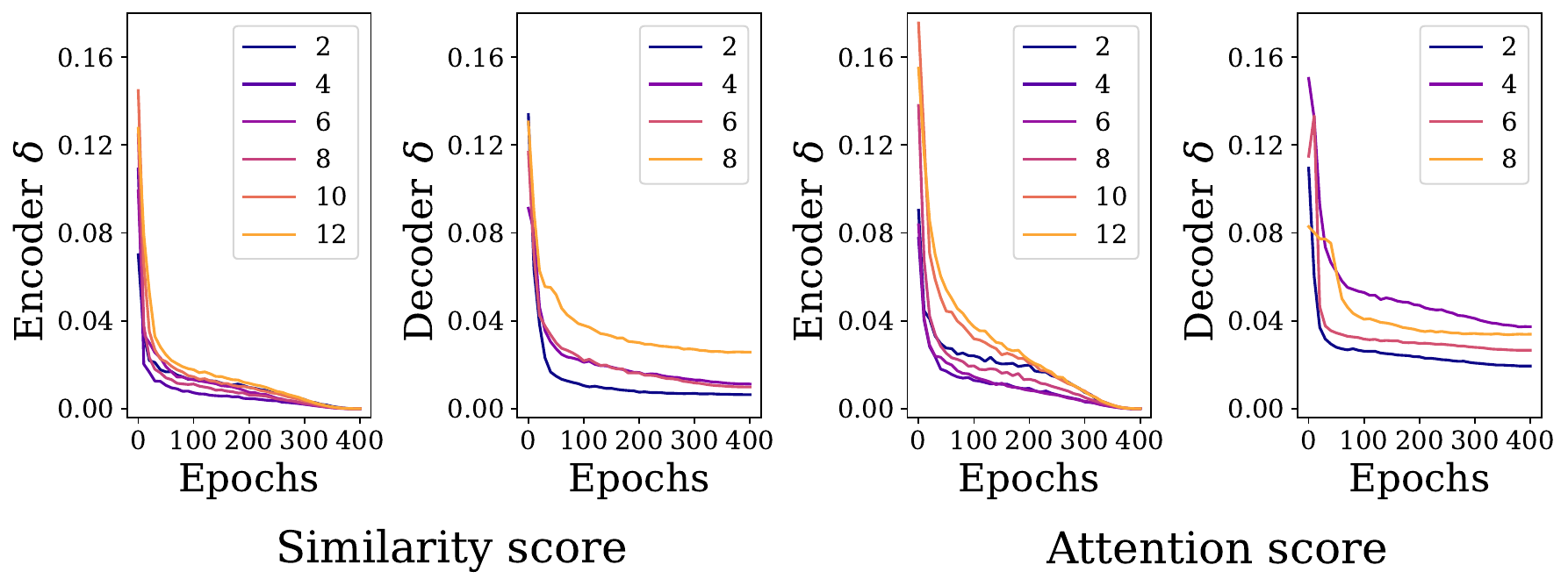}
        \caption*{(b) KL divergence of token relations}
    \vspace{-0.0cm}
    \caption{\textbf{MAE learns patch clustering from very early stage of training process.} (a) MAE widens the gap $\mu_\text{intra} - \mu_\text{inter}$. (b) Token relations drastically converge at early epochs and then gradually level off. Numbers in the legend denote the layer $i$.
    More details are provided in
    \cref{appendix:convergence}.}
    \label{fig:bipartition_kld}
\end{figure}



%% file: sec/figure_latex/mask_ratio_new.tex

\begin{wrapfigure}{r}{0.4\textwidth}
    \vspace{-.64cm}
            \begin{center}
    \hspace{-.04\textwidth}
            \includegraphics[height=0.16\textheight]{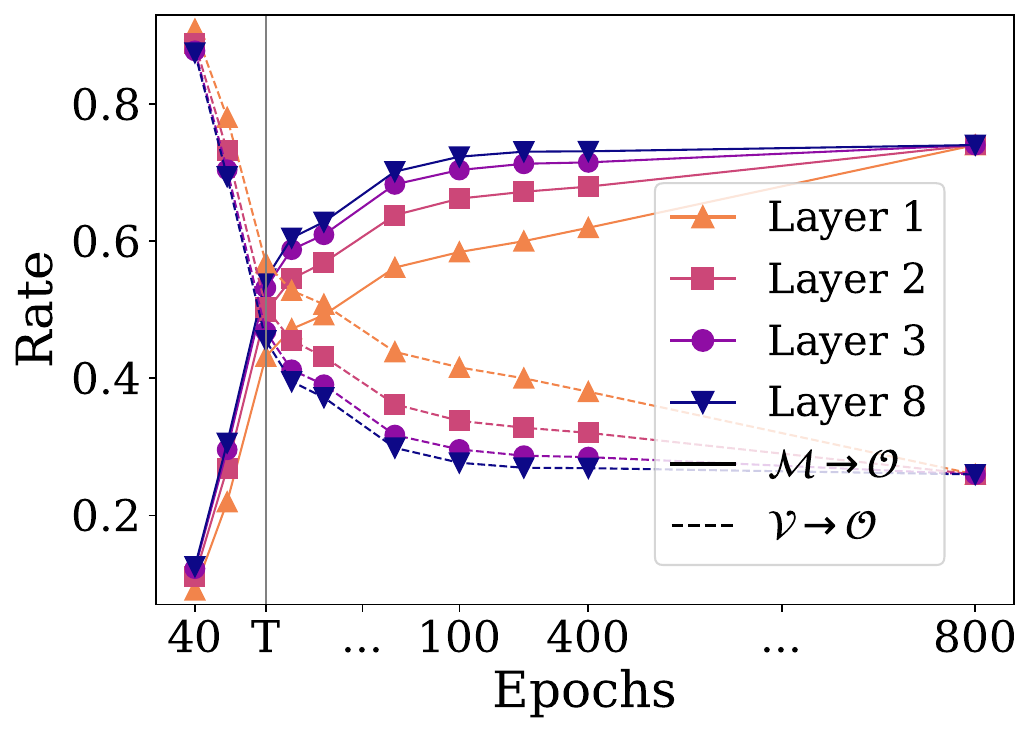}
            \end{center}
            \vspace{-0.4cm}
        \caption{\textbf{Exploitation rate.}}
        \label{fig:mask_ratio_new}
    \vspace{-0.7cm}
\end{wrapfigure}


%% file: sec/4_approach.tex
\section{Self-Guided Informed Masking}
\label{sec:approach}


In \cref{sec:analysis:MAE_clustering} and \cref{sec:analysis:early}, we show that the MAE encoder learns patch clustering from an early stage, allowing us to appropriately bi-partition the image into two major token clusters and to mask out one of them.
In other words, we can generate informed masks with MAE itself early in the pre-training phase and use these informed masks for the remainder of the training.
To decide when exactly the MAE can properly cluster the patches, we use exploitation rate suggested in \cref{sec:analysis:decoder}, which allows us to confidently generate informed masks at $T$ epoch, ultimately leading to the design of our method.

From these observations in \cref{sec:analysis}, we are motivated to leverage the patch relevance learned from the early-stage 
to expedite training, instead of relying on random masking.
Random masking delays the learning of powerful patch-clustering, inefficiently revisiting easily separable patches already clustered in the early stage which reflects the \emph{key dissimilarities} among the image tokens.

Based on this idea, we propose our \emph{self-guided informed masking}, which internally injects the information about the learned \emph{key dissimilarities} by intensively masking one of the top two well-separated clusters. 
We emphasize that MAE is still trained in a single stage; at epoch $T$, we begin generating informed masks and continue the training process without interruption.

Armed with our method, we can accelerate MAE to focus on learning less distinguishable patches instead of wasting time repeating to discover the most prominent patterns. As our method purely relies on inherent metrics during training, it is \textit{completely free from any external models or extra information}. A more detailed reasoning can be found in \cref{appendix:reasoning}.

To achieve this, we need to 1) bi-partition the image, 2) properly design the informed masks, 3) select the attention layer to construct the informed masks, and 4) decide when to start informed masking.

\textbf{Bi-partitioning.}
To bi-partition the image reflecting the learned \emph{key dissimilarities}, we take Normalized Cut (Ncut)~\cite{shi2000normalized} to consider both dissimilarity between different clusters and similarity inside each cluster.
We construct a fully-connected undirected image graph with $\mathbf{M}$ (\cref{eq:cosine_similarity}) which consists of the patches
and similarity between
them 
as nodes and edges, respectively. 
To partition the set of all node indices $\mathcal{O}$
into two disjoint sets $\mathcal{A}$ and $\mathcal{B}$, we minimize the Ncut energy $S(\mathcal{A}, \mathcal{B})/S(\mathcal{A}, \mathcal{O}) + S(\mathcal{A}, \mathcal{B})/S(\mathcal{B}, \mathcal{O})$, 
where $S(\mathcal{A}, \mathcal{B}) = \sum _{{i} \in \mathcal{A}, {j} \in \mathcal{B}} \mathbf{M}_{i,j}$.
As shown in~\cite{shi2000normalized}, we can approximate the solution of this problem by calculating the second smallest eigenvector ($\mathbf{y}_{1}$) of the eigensystem $(\mathbf{D} - \mathbf{M})\mathbf{y} = \lambda \mathbf{D}\mathbf{y}$, where $\mathbf{D}$ is a diagonal matrix with $\mathbf{D}_{i,i} = \sum _{j}\mathbf{M}_{i,j}$ and $\mathbf{D} - \mathbf{M}$ is the Laplacian matrix. Finally, we bi-partition the graph by thresholding $\mathbf{y}_{1}$ with its mean, $\overline{\mathbf{y}_{1}}$, \emph{i.e.}, 
$\mathcal{A} = \{{i}|\mathbf{y}_{1}^{i} \geq \overline{\mathbf{y}_{1}}\}$ and $\mathcal{B} = \{{i}|\mathbf{y}_{1}^{i} < \overline{\mathbf{y}_{1}}\}$.

\input{sec/figure_latex/mask_example}
\textbf{Object-centric Masking.}
Our approach stems from the observation that masking the entire image leads to learn patch clustering across the entire image; \textit{i.e.}, the reconstruction loss affects the whole image.
To refine this process, we restrict the masking to object-centric regions.
By narrowing the masking focus, our method guides the MAE to concentrate on learning patch clustering within the object regions; that is, the loss affects only the object-related parts, thereby accelerating the process of learning patch clustering in object region. In this context,
we aim to mask out the cluster $\mathcal{C}$ containing the main object, letting the model to learn feature representations for the foreground faster.
Since we do not have access to the label, we take an indirect approach:
the token with the largest absolute element in $\mathbf{y}_{1}$ tends to compose the main object.~\cite{wang2022tokencut}

In reality, however, practical issues like imperfect bi-partitioning and varying cluster sizes complicating batch processing, arise.
To address,
we rank the tokens by the relevance to $\mathcal{C}$ and mask out a fixed ratio of the tokens based on the ranking.
The \emph{relevance score} $S_i$ of the $i$-th patch is defined as
\begin{equation}
  S_i = \frac{{\bar{\mathbf{x}}_m}^\top \mathbf{x}'^{(i)}}{{\| \bar{\mathbf{x}}_m \|_2} \cdot {\|\mathbf{x}'^{(i)} \|_2}},
  \quad \text{where} \ \
  \bar{\mathbf{x}}_m = \frac{1}{|\mathcal{C}|} \sum_{i \in \mathcal{C}} \mathbf{x}'^{(i)}.
  \label{eq:si_score}
\end{equation}
As $\bar{\mathbf{x}}_m$ represents the majority of $\mathcal{C}$ which mainly consists of the object, $S_i$ exhibits the relevance to the object by measuring the similarity of each patch to $\bar{\mathbf{x}}_m$.
In this manner, we can robustly mask out the whole object even with the noisy bi-partitioning.


Though intensive masking on an object leads to expedited feature learning on it, 
reconstruction is fundamentally impossible if all tokens are masked,
as there is no clue to construct high-level information about it.
To prevent such a case, we add a few hint tokens, chosen either by uniformly randomly or proportional to $S_i$ with decaying ratio, following~\cite{li2022semmae}.
\cref{fig:mask_examples} illustrates our object-centric informed masking (4th row) based on bi-partitioning (3rd row).
We provide minimum information for estimating shared information via the hint tokens (last row).




\input{sec/figure_latex/attn_nmi}
\textbf{Appropriate Layer for Patch Clustering.}
We consider attention distance and Normalized Mutual Information (NMI) in \cref{fig:attn_nmi}, to decide with which layer embeddings to compute the similarity matrix among the patches.
To obtain sufficiently meaningful token relation, we discard the early layers (\textasciitilde4th layer) from the candidates as they show extremely low NMI, which indicates relatively homogeneous attention map.
Then, we select the second last encoder layer with the highest attention distance, as they have the highest potential awareness of the global patterns.



\textbf{When to Start Informed Masking.}
Lastly, we need to determine when the model has learned patch clustering enough to generate high-quality informed masks. Since the \emph{relative} relationship among the tokens is enough for bi-partitioning,
we generate them when the mask tokens start to have comparable amount of information to the visible tokens.
Based on \cref{sec:analysis:decoder}, we quantify the
shared information possessed by the mask tokens and start informed masking when it becomes comparable to the information in visible tokens ($R^{(L)}_{\mathcal{M} \rightarrow \mathcal{O}} \geq R^{(L)}_{\mathcal{V} \rightarrow \mathcal{O}}$), \emph{i.e.}, around $T$ epochs in \cref{fig:mask_ratio_new}.

%% file: sec/figure_latex/mask_example.tex
\begin{figure}[t]
    \centering
    \includegraphics[width=1.\linewidth]{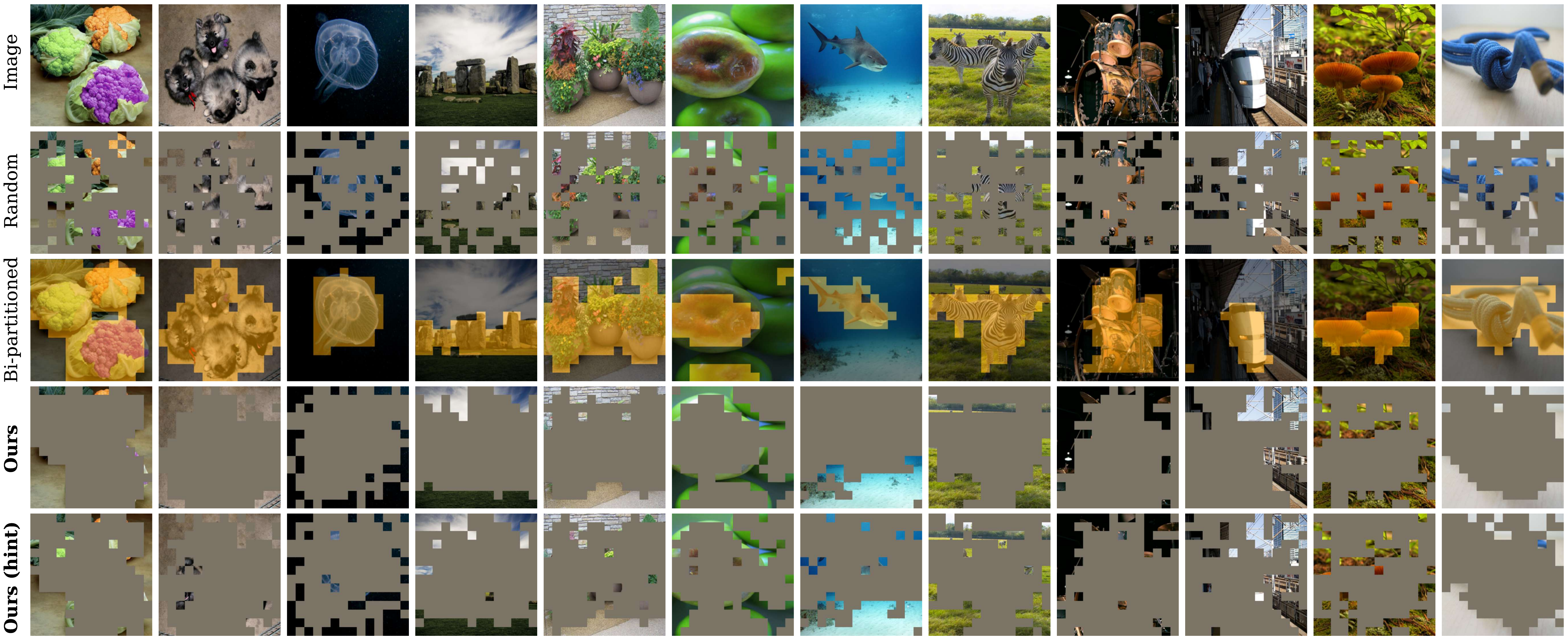}
    \caption{
        \textbf{Examples of Self-guided Informed masking.} 
        More examples and detailed explanations on our method are displayed in \cref{appendix:reasoning}.
    }
    \label{fig:mask_examples}
\end{figure}

%% file: sec/figure_latex/attn_nmi.tex
\begin{wrapfigure}{r}{0.4\textwidth}
    \vspace{-.65cm}
            \begin{center}
            \hspace{-0.2cm}
        \includegraphics[height=0.15\textheight]{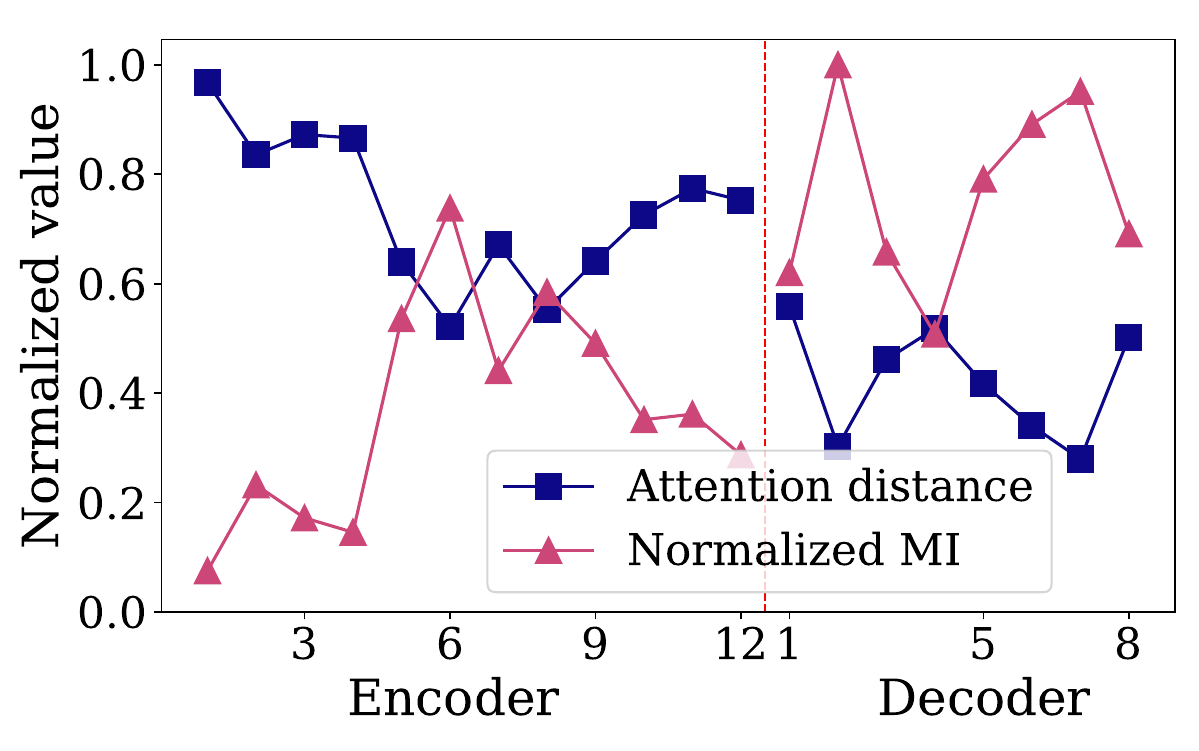}
            \end{center}
            \vspace{-0.3cm}
        \caption{\textbf{MAE properties.}}
        \label{fig:attn_nmi}
    \vspace{-0.4cm}
\end{wrapfigure}

%% file: sec/5_experiments.tex
\section{Experiments}
\label{sec:exp}

\subsection{Experimental Settings}
\label{sec:exp:setting}

\noindent
\textbf{Baselines.}
We compare our model to the original MAE~\cite{he2022masked} and Attention-driven Masking and Throwing (AMT)~\cite{liu2023amt}, a recently enhanced MAE without requiring any external model or label information. 
We exclude other models requiring external pre-trained models~\cite{chen2023automae, li2022semmae} or labeled data~\cite{kakogeorgiou2022hide} for generating informed masks, since it is no longer a fair comparison.

\textbf{Experimental Protocol.}
We pre-train all competing models for 400 epochs on ImageNet-1K \cite{deng2009imagenet}, and fine-tune on 3 downstream tasks: image classification, object detection, and semantic segmentation.
All experiments are conducted following the settings in original MAE~\cite{he2022masked}, unless noted otherwise.
We conduct experiments on 8 NVidia A6000 GPUs (48GB).


\textbf{Datasets.}
We use CIFAR-100~\cite{krizhevsky2009cifar}, iNaturalist 2019~\cite{van2018inaturalist}, and CUB200-2011~\cite{wah2011cub} for image classification.
We fine-tune our model on COCO~\cite{lin2014coco} for object detection, and on ADE20K~\cite{zhou2019ade20k} for semantic segmentation.


\subsection{Performance on Downstream Tasks}
\label{sec:exp:transfer}

\noindent \textbf{Image Classification.}
The left side of \cref{tab:downstream_tasks} compares the image classification performance on various datasets.
Our method outperforms the baselines on all datasets with both linear probing and fine-tuning, implying that the expedited training with our method leads to stronger feature representation after same epochs of training.
We provide further analysis on our boosted performance in \cref{sec:exp:result} and extended training results in \cref{sec:exp:scalability}.

\textbf{Object Detection and Segmentation.}
We fine-tune a Mask R-CNN model~\cite{he2017rcnn} end-to-end on COCO with a ViT backbone for 90K iterations, and evaluate with the average precision for bounding box ($\text{AP}^\text{box}$) and for segmentation mask ($\text{AP}^\text{mask}$).
We also compare semantic segmentation performance on ADE20K using UperNet~\cite{xiao2018unified}, in Mean Intersection over Union (mIoU). 
Our method outperforms baselines in all fine-grained tasks as shown in the right side of \cref{tab:downstream_tasks}, indicating that ours better captures the fine details of the image with the same training session.
\input{sec/table_latex/downstream_tasks}

\subsection{Ablation Studies}
\label{sec:exp:ablation}

\cref{tab:ablation} compares linear-probing performance of our method on image classification, with various settings.
We fix the masking ratio to 0.75 for all the experiments in this section. Ablation studies on more factors including masking ratio can be found in \cref{appendix:ablation}.
The first group verifies that the later layers of the encoder yield the most accurate token relations, aligned well with our analysis in \cref{sec:approach}.
The second group verifies that the hint tokens proposed in \cref{sec:approach} are essential.
Also, better performance with uniform sampling (Random) than with $S_i$-based approach (\cref{eq:si_score}) indicates the importance of providing equal opportunity to all clusters to have visible tokens.
\input{sec/table_latex/ablation}

\subsection{Analysis on the Learned Feature Space}
\label{sec:exp:result}

We take a deeper look into our method for further insights on its improvements via various metrics with the ImageNet-1K validation set.
We analyze with $m = 0$ unless noted otherwise.

\textbf{Attention Distance~\cite{dosovitskiy2020image}.}
We measure the weighted average distance of the attention operations between the query and key tokens within the image in Figure \hyperref[fig:analysis]{\ref*{fig:analysis}a}.
Since it can be interpreted as the size of the receptive fields in CNNs, higher attention distance of our method suggests that it has been better-trained at the same epoch, more globally capturing the image context.


\textbf{Fourier Analysis~\cite{park2023selfvit}.}
Figure \hyperref[fig:analysis]{\ref*{fig:analysis}b} shows the relative log amplitude of Fourier-transformed representations, which indicates the degree to which the model prioritizes either high-frequency (pattern)
or low-frequency (shape) information.
Our method utilizes high-frequency components more intensively than MAE does,
implying more powerful pattern-based clustering.

\textbf{Mask Token Variance.}
We report the variance of mask token embeddings along the decoder layers with $m = 0.75$ in Figure \hyperref[fig:analysis]{\ref*{fig:analysis}c}.
As they carry high-level semantics of each potential patch cluster, higher variance among them indicates that the latent variables responsible for estimated shared information in each individual cluster has been diversified, implying that patches are grouped into finer clusters.
Consistently higher variance along the layers of ours manifests its further progressed patch clustering.


\textbf{Qualitative Comparison.}
\cref{fig:intro} clearly shows that ours captures finer patch embeddings and tighter boundaries than MAE. See \cref{appendix:qualitative} for more examples and detailed explanations.




%% file: sec/table_latex/downstream_tasks.tex
\begin{table}[t]
    \centering
    \captionsetup{width=1.\linewidth}
    \caption{\textbf{Performance on downstream tasks.} LP and FT stand for Linear probing and Fine-tuning, respectively. Det. indicates the Object Detection task.}
    \vspace{0.2cm}
    \footnotesize
    \renewcommand{\tabcolsep}{4pt}
    \resizebox{1.\linewidth}{!}{
    \hspace{-0.13cm}
    \begin{tabular}{l|C{1.2cm}C{1.2cm}C{1.2cm}C{1.2cm}C{1.2cm}|C{1.2cm}|C{1.2cm}C{1.3cm}} 
        \toprule
        Task & \multicolumn{5}{c|}{Image Classification} & Det. & \multicolumn{2}{c}{Segmentation} \\
        Dataset~~ & \multicolumn{2}{c}{ImageNet-1K} & iNat2019 & CIFAR & CUB & COCO & COCO & ADE20K \\
        Metric & LP & FT & FT & FT & FT & $\text{AP}^\text{box}$ & $\text{AP}^\text{mask}$ & mIoU \\
        \midrule
        MAE~\cite{he2022masked} & 61.4 & 82.5 & 78.7 & 89.3 & 81.8 & 43.0 & 38.9 & 45.0 \\ 
        AMT~\cite{liu2023amt} & 61.7 & 82.8 & 76.0 & 87.8 & 80.8 & 42.8 & 36.6 & 43.1 \\ 
        Ours & \textbf{62.9} & \textbf{83.2} & \textbf{78.9} & \textbf{90.0} & \textbf{82.8} & \textbf{43.3} & \textbf{39.3} & \textbf{45.2} \\ 
        \bottomrule
    \end{tabular}
    }
    \label{tab:downstream_tasks}
\end{table}


%% file: sec/table_latex/ablation.tex
\newcolumntype{Y}{>{\centering\arraybackslash}X}
\definecolor{lightgray}{gray}{0.9}
\begin{table}[h]
    \centering
    \captionsetup{width=.8\linewidth}
    \caption{\textbf{Ablation studies.} The default is highlighted in gray.
    Detailed analysis can be found in \cref{appendix:ablation}.}
    \vspace{0.2cm}
    

    \footnotesize  
    \setlength{\tabcolsep}{4pt}  
    \begin{tabular}{>{\centering\arraybackslash}p{1.6cm} >{\centering\arraybackslash}p{2.6cm} >{\centering\arraybackslash}p{2.6cm} >{\centering\arraybackslash}p{2.6cm}}
    \toprule
    Layer & Target cluster & Hint strategy & Linear probing \\
    \midrule
    Enc 3 & Object & Random & 62.3 \\
    Enc 7 & Object & Random & 62.4 \\
    Dec 8 & Object & Random & 62.7 \\ 
    \midrule 
    Enc 11 & Object & $S_i$-based & 62.5 \\
    Enc 11 & Object & No hint &  52.3\\
    \midrule
    \rowcolor{lightgray}
    Enc 11 & Object & Random &  \textbf{62.9}\\
    \bottomrule
\end{tabular}

    \label{tab:ablation}
\end{table}


%% file: sec/6_related_work.tex
\section{Related Work}
\label{sec:related_work}
\input{sec/figure_latex/attn_dist_nmi_fourier}

\noindent \textbf{Masked Image Modeling (MIM).}
Inspired by Masked Language Modeling~\cite{devlin2018bert,brown2020language}, 
MIM has been widely applied in image~\cite{zhou2021ibot, chen2020iGPT, pathak2016context, bao2021beit, xie2022simmim, he2022masked, dong2023peco, atito2021sit, assran2022siamese, xie2022mfm, girdhar2023omnimae, li2022mc, atito2021mc, gao2022mcmae, liu2023pixmim, wang2023masked, fang2023unleashing, bachmann2022multimae, liang2022supmae} and video understanding~\cite{tong2022videomae, wang2023videomae, feichtenhofer2022masked, wang2022bevt, mun2022bassl, bandara2023adamae, wei2022maskfeat, yan2021videogpt, wang2022less, gupta2022maskvit, gupta2023siamese, huang2023mgmae, huang2022contrastive, hou2022milan}.
Context Encoder~\cite{pathak2016context} established MIM with CNNs, focusing on the prediction of masked regions of an image.
MLM also has been applied to ViT~\cite{dosovitskiy2020image}; \emph{e.g.}, BEiT~\cite{bao2021beit} leverages visual tokens from dVAE as reconstruction targets for masked image patches, and SimMIM~\cite{xie2022simmim} directly predicts raw pixel values as a regression problem. 
MAE~\cite{he2022masked} also regresses the raw pixels, adopting asymmetric encoder-decoder architecture.



\vspace{0.1cm} \noindent \textbf{Informed Masking.}
Recent researches have considered to arm MAE with an advanced masking strategy~\cite{li2021mst, kakogeorgiou2022hide, shi2022adversarial, chen2023automae, li2022semmae, liu2023amt}.
MST~\cite{li2021mst} and AttMask~\cite{kakogeorgiou2022hide} have pioneered information-guided masks, utilizing the attention maps of a supervised ViT.
ADIOS~\cite{shi2022adversarial} adopts adversarial training to get optimal masks.
SemMAE~\cite{li2022semmae} and AutoMAE~\cite{chen2023automae} leverage the powerful knowledge from pre-trained self-supervised ViTs, showcasing the synergistic fusion of informed masking with MAE.
Despite effectiveness, these methods are limited as they require an external model or rely on labels, which make these models no longer fully self-supervised.
To address this critical issue, AMT~\cite{liu2023amt} extracts attention maps directly from the model during pre-training and generates informed masks from them.

\vspace{0.1cm} \noindent \textbf{Analysis on MAE.}
Nam \emph{et al.}~\cite{park2023selfvit} reports a comparative analysis between MIM and contrastive learning,
highlighting how the encoder-decoder architecture of MAE empowers the properties of MIM, although their discovery is confined to token differentiation of MIM models.
Another avenues introduce theoretical architectures to elucidate MAE's behavior under specific assumptions~\cite{cao2022understand,pan2022towards,zhang2022mask, lee2021predicting}.
Hierarchical latent variable model~\cite{kong2023hierarchical} aligns well with our primary observations.

%% file: sec/figure_latex/attn_dist_nmi_fourier.tex
\begin{figure*}[t]
    \centering
    \hspace{-0.5cm}
    \includegraphics[height=0.135\textheight]{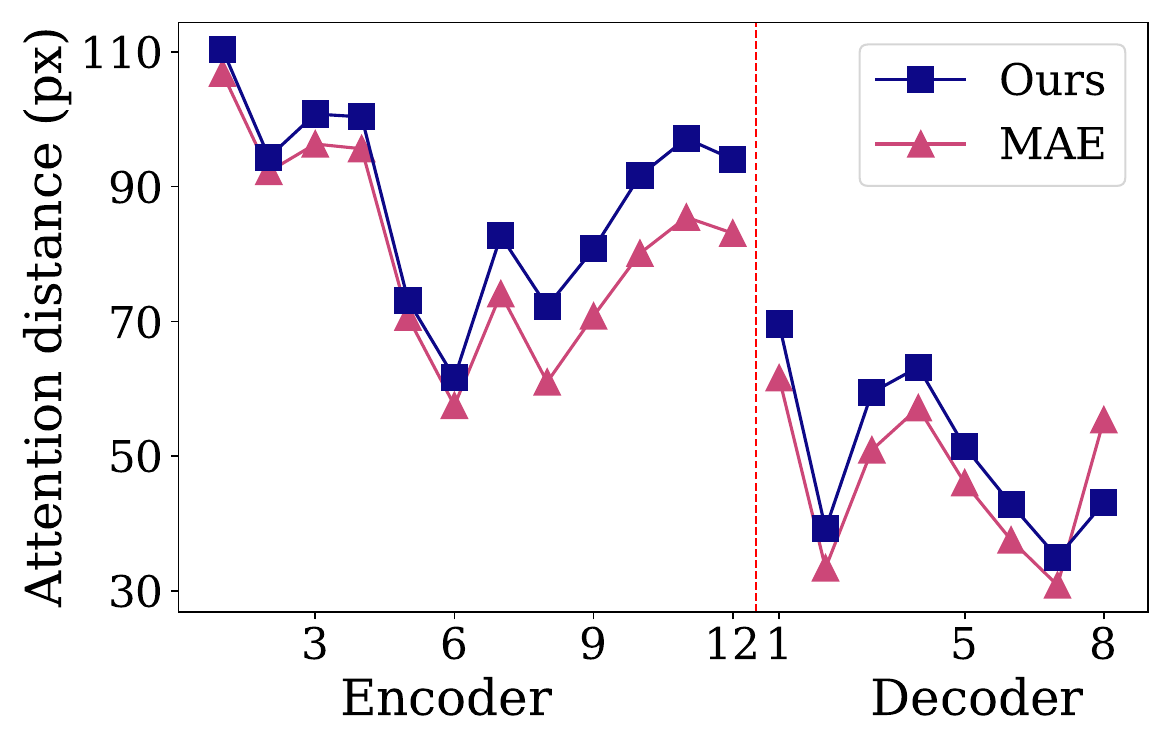}
    \hspace{.5cm}
    \includegraphics[height=0.135\textheight]{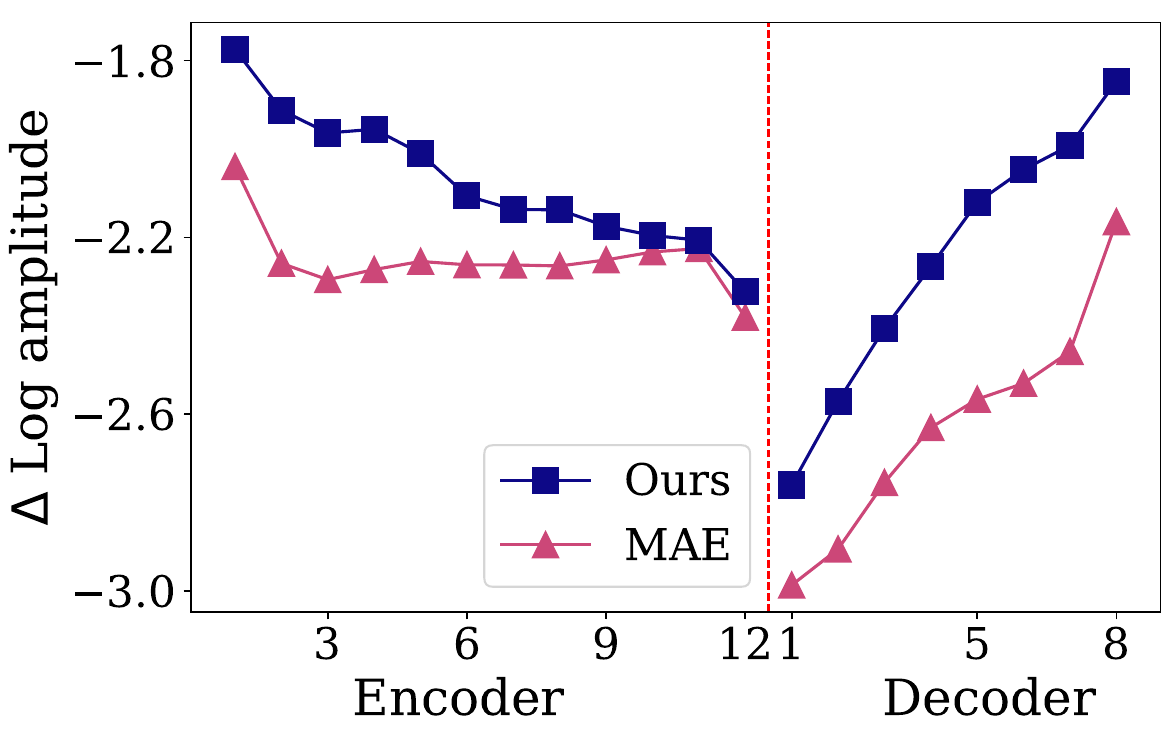}
    \hspace{.5cm}
    \includegraphics[height=0.135\textheight]{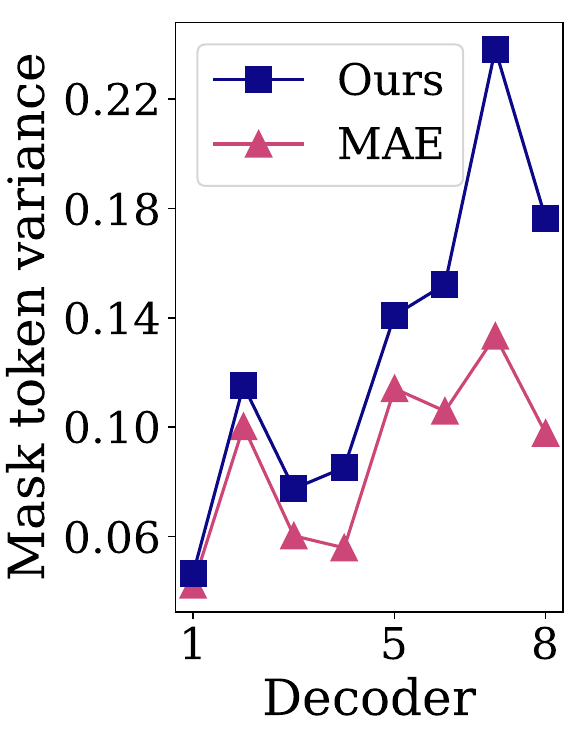}
    \caption*{~~~~~~~~~~~~(a) Attention distance~~~~~~~~~~~~~~~~~~~~~~~~~~~~~~(b) Fourier analysis~~~~~~~~~~~~~~~~(c) Mask token variance}
    \vspace{-0.0cm}
    \caption{\textbf{Metrics explaining our performance gain.} Layers left on the red dotted line belong to the encoder, and the rest to the decoder.
    }
    \label{fig:analysis}
\end{figure*}



%% file: sec/7_conclusion.tex
\section{Summary and Limitations}
\label{sec:conclusion}

Unveiling the operation of Masked Autoencoder (MAE) which fundamentally learns pattern-based patch-level clustering, we expedite the MAE to learn patch clustering by incorporating informed mask derived from itself.
Notably, our method does not require any other external models or additional information.
Superior results on extensive experiments demonstrate the effectiveness of our method.

\textbf{Limitations.}
Our method may show less significant improvement when training with excessively fragmented images, \emph{e.g.,} some dataset for segmentation tasks.
In detail, since there would be numerous clusters within each image, masking specific clusters with informed masking may yield similar masks to random masking.

%% file: X_suppl.tex
\clearpage

\setcounter{page}{1}

\appendix

\pagenumbering{roman}
\renewcommand\thetable{\Roman{table}}
\renewcommand\thefigure{\Roman{figure}}
\setcounter{table}{0}
\setcounter{figure}{0}

\crefname{table}{Table}{Tables}
\crefname{figure}{Figure}{Figures}
\crefname{section}{Section}{Sections}
\crefname{appendix}{Appendix}{Appendices}
\crefname{equation}{Equation}{Equations}
\section*{Appendix}

\section{Method Elaboration}
\label{appendix:reasoning}
\textbf{Detailed Reasoning for Our Method.}
As discussed in \cref{sec:preliminary}, the true shared information $\mathbf{c}$ exists for the entire token set $\mathcal{X}$, which is equivalent to statistical dependency among the patches in $\mathcal{X}$.
With training, MAE learns to estimate this high-level latent variable $\hat{\mathbf{c}}$, which reflects the context of the entire image.
Let us denote by $\mathbf{s}_m$ and $\mathbf{s}_v$ for information specific to masked out patches $\mathcal{X}_m$ and visible patches $\mathcal{X}_v$ respectively, \emph{e.g.}, positional embeddings.

Since MAE cannot access $\mathcal{X}_m$ during training, the decoder is forced to reconstruct $\mathcal{X}_m$ via 1) simple interpolation using visible tokens, or 2) estimated statistical dependency among the entire tokens, \emph{i.e.}, $\hat{\mathbf{c}}$.
As shown in \cref{fig:hierarchical_appendix}, simple interpolation means reconstructing $\mathcal{X}_m$ mainly with $\mathcal{X}_v$ and $\mathbf{s}_v$, which is not directly related to $\mathcal{X}_m$, leading to poor reconstruction result.
However, due to the reconstruction loss, MAE is forced to improve the reconstruction quality, establishing high-level information $\hat{\mathbf{c}}$ and performing the reconstruction based on it.
As a result, at some moment, the encoder starts to map the visible tokens $\mathcal{X}_v$ to estimated shared information $\hat{\mathbf{c}}$ for the whole token set $\mathcal{X}$, and decoder exploits this hierarchical information to reconstruct the low-level information; \emph{i.e.}, the raw RGB pixels of $\mathcal{X}_m$.
This process is verified in \cref{fig:mask_ratio_new} in the main manuscript.

\input{sec/figure_latex/hierarchical_appendix}

Moreover, connecting this logic to our discovery in \cref{sec:analysis:MAE_clustering}, we claim that this unknown $\mathbf{c}$ conceptually corresponds to pattern-based patch clustering information. In other words, considering the pattern-based patch clustering in MAE (as verified in \cref{sec:analysis}), it suggests that MAE clusters the patches and builds corresponding high-level variable containing $\hat{\mathbf{c}}$ for \emph{each} patch cluster.

In summary, MAE learns to construct the latent variables for each potential patch cluster. However, considering the fact that MAE learns \emph{relevance} among the patches from the extremely early stages in pre-training process (\cref{sec:analysis:early}), it can be inferred that MAE with naive random masking is actually revisiting \emph{key dissimilarities} in $\mathcal{X}$, which exists between easily separable patches, every epoch wasting large portion of its training resources. Especially, when it comes to bi-partitioning (which is the simplest form of \emph{key dissimilarities}), MAE learns it from the very early epochs as verified in Fig. \newhref{fig:bipartition_kld}{a}.

Based on this reasoning, we can enforce MAE to focus on learning hardly distinguishable tokens by guiding MAE to skip revisiting \emph{key dissimilarities} by injecting the information about it as input. We can inject this information via informed masks, which possess \emph{key dissimilarities} by intensively masking one of the bi-partitioned clusters, leading MAE to assign most of the training resource to learning relatively vague patch clusters in masked out patch sets.

\input{sec/figure_latex/appendix_mask}
\textbf{Qualitative Analysis.}
As discussed in \cref{sec:approach}, our method generates informed masks by itself without using any external model or requiring additional information.
Recall that MAE generates informed masks after $T \approx 50$ epochs of training.
\cref{fig:appendix_mask} compares our informed masking with and without the hint tokens to the random masking.
It also illustrates the bi-partitioned clusters extracted from MAE itself after 51 epochs, which are used for the internal generation of the informed masks.
We observe in these examples that our relevance-score-based masking (\cref{sec:approach}) guarantees to fully mask out the target cluster even when the bi-partitioning is not perfect.
For example, the target cluster in (e) consists of the portion of \emph{house} and \emph{sky}, but our method fully masks out the patches composing the \emph{house} in the image.
Similar results can be found in (j) and (k).
Also, even when the foreground is not clearly distinguished due to the barely discernible patterns as in (i) and (l), we see that our approach still fully masks out the object.
The success of relevance score strongly indicates that patch vectors are hierarchically clustered based on their \emph{visual patterns}, as they are masked out in the \emph{order} of pattern similarity with the mean patch vector.

We confirm from the examples that even in early epochs, MAE is able to appropriately bi-partition the image, which means it has already learned to discriminate the image into two clusters.
We also find that most of the examples are bi-partitioned into foreground and background, since the similarity edges between these two groups tend to have the weakest values.
In summary, although MAE in the early epochs does not promise to provide perfectly discriminated object-centric cluster from the image, our proposed approach robustly builds object-centric masks through the introduction of the relevance score.


\section{Token Relations}
\label{appendix:convergence}

\noindent \textbf{Patch Clustering in Projected Latent Space.}
\cref{fig:mae_tsne} illustrates the patch clusters on a few examples and their t-sne plots.
We consider a graph $\mathcal{G} = (\mathcal{V}, \mathcal{E})$ for the given image, where $\mathcal{V}$ and $\mathcal{E}$ correspond to patches and edges between them weighted by $\mathbf{M}$ in Eq.~\eqref{eq:cosine_similarity}, respectively.
From this graph, we repeatedly apply Normalized Cut~\cite{shi2000normalized} to remove edges with the lowest relevance until the graph is split into a predefined number ($K$) of clusters.
We clearly see that tokens with similar visual patterns (color, texture) are 1) grouped together as the same patch cluster (2nd row) and 2) embedded closely in the latent space (last row). Apparent discrimination in the representation space supports the patch-level clustering.
\input{sec/figure_latex/mae_tsne}

\input{sec/table_latex/feat_var_appendix}

\noindent \textbf{Enhanced Feature Separability of MAE with Our Method.}
Based on our analysis on embedding space suggested in \cref{eq:feature_sim_variance}, we compare the vanilla MAE and our method in the aspect of feature separability with 800 epochs of training in \cref{tab:variance_rebuttal}.
The results indicate that our method shows more diversified feature space via higher feature variance and similarity variance, aligned well with the analysis in \cref{sec:exp:result}.

\noindent \textbf{KL Divergence of All Layers in MAE.}
We additionally provide KL divergence (KLD) of token relations for all layers in MAE as an extension of \cref{sec:analysis:early}.
For the decoder, we use token relations with the intact input, \emph{i.e.}, $D([E(\mathbf{X})])$, for the criterion distribution in the KLD (\cref{eq:kld}).
In other words, we compare the token relations from each epoch with masked inputs to the token relations from the last epoch with intact inputs.
Due to this setting, KLD with decoder does not converge to zero at the final epoch in \cref{fig:appendix_kld}.

As shown in \cref{fig:appendix_kld}, all layers but the first one in the decoder drastically converge at the early epochs with both of $\mathbf{M}$ than $\mathbf{A}$.
Encoder ($E(\mathbf{X})$) layers are much stabler and converge faster than decoder ($D([E(\mathbf{X}_v);\mathbf{m}])$) layers due to the difference in the amount of given information.
Also, since the cosine similarity scores $\mathbf{M}$ directly compare the similarity among the tokens, strong convergence of $\mathbf{M}$ supports the observation that MAE intrinsically learns the patch-level clustering.

KLD of the attention scores in the first encoder layer is low at the first epoch, which implies that it learns homogeneous attention map rather than random values as discussed in \cref{sec:exp:result}.
The first layer of the decoder shows high KLD with the attention scores along with the training, because 1) the mask tokens are not contextualized yet 
(that is, mask token vectors does not represent the masked out patches at all),
and 2) the index of each mask token is randomly selected for every epoch.
On the other hand, KLD with the similarity scores decreases along the epochs, because the similarity score matrix is calculated after the contextualization.
This suggests that even a single first layer in decoder has ability to properly exploit $\hat{\mathbf{c}}$ from the encoder to discriminate the patches although it is weaker than the later layers.
\input{sec/figure_latex/appendix_kld}

\input{sec/figure_latex/appendix_bipartitionings}
\textbf{Further Experiments on ViT~\cite{dosovitskiy2020image} and MoCo~\cite{he2020moco}.}
We provide bi-partitioning performance and KL divergence of token relations of ViT and MoCo for better understanding on our metrics in \cref{fig:appendix:bipartitioning}. We display the result of MAE encoder together for comparison. Before delving into the analysis, we note that the result of this experiment with ViT and MoCo is irrelevant to our main claims since ViT and MoCo do not learn patch clustering.

As MoCo yields homogeneous attention map~\cite{park2023selfvit} resulting in simple form of embedding space, \emph{e.g.}, main object cluster and background cluster, 
the result of MoCo in \cref{fig:appendix:bipartitioning} indicates that the last epoch of MoCo has provided properly bi-partitioned patch groups. Consistent gap between mean inter-cluster ($\mu_\text{inter}$) and mean intra-cluster ($\mu_\text{intra}$) edge weights of similarity score matrix $\mathbf{M}$ and attention score matrix $\mathbf{A}$ of MoCo supports this claim.

Unlike MAE or MoCo, embedding space of ViT does not guarantee to provide appropriate bi-partitioning results.
As a result, in \cref{fig:appendix:bipartitioning}, although the similarity score matrix $\mathbf{M}$ enlarges the gap between $\mu_\text{inter}$ and $\mu_\text{intra}$, the attention score matrix $\mathbf{A}$ increases the $\mu_\text{inter}$ rather than $\mu_\text{intra}$. This hardly interpretable pattern implies that the pseudo-ground truth for bi-partitioned patch groups generated at the last epoch is unstable or even incorrect.

In summary, only MAE explicitly shows its ability to clearly recognize \emph{key dissimilarities} among the tokens, \emph{i.e.}, bi-partitioning information, from the extremely early stage of pre-training, and consistently escalates the gap between $\mu_\text{inter}$ and $\mu_\text{intra}$.

\input{sec/figure_latex/appendix_kld_moco_vit}
\cref{fig:appendix:kld_moco_vit} shows the KL divergence of token relations from ViT and MoCo. Compared to the result of MAE in \cref{fig:appendix_kld}, both ViT and MoCo reveal gradual convergence of token relations and some layers exhibit their unstable convergence. Again, as ViT and MoCo do not learn patch-clustering, the experiment results of ViT and MoCo are off-topic to the main stream of our work.

\input{sec/figure_latex/appendix_qualitative}
\section{Qualitative Results}
\label{appendix:qualitative}

We provide more qualitative examples of patch clustering compared to vanilla MAE in \cref{fig:appendix_qual}, where we see that images are segmented into $K$ clusters in unsupervised manner.
Successful segmentation from our recursive graph-cut suggests that features are hierarchically discriminated in the embedding space.
Our method clearly shows more accurately clustered patches based on their pattern and also yields tighter boundary between the clusters for various types of images, \emph{i.e.}, object-centered images and those containing higher portion of background.

\section{Analysis on Ablation Studies}
\label{appendix:ablation}

As displayed in \cref{tab:ablation_full}, our ablation study on layer selection for embedding extraction verifies the hypothesis on it (See \cref{sec:analysis}), while showing the minor effect on model performance relative to other factors.
Especially, 
the last layer of the decoder shows higher performance than the early or intermediate layers of the encoder.
Since the decoder possesses the patch cluster information constructed through the entire encoder layers, it may have more appropriate bi-partitioning quality than using a few early encoder layers, \emph{e.g.}, layer 3 or layer 7.

To analyze the reason for the minor effect on layer selection, we display the examples of informed masks generated with bi-partitioned patch cluster from each layer in \cref{fig:appendix_layer}. 

\input{sec/table_latex/ablation_full_appendix}

In \cref{fig:appendix_layer}, we find that the later layer of the encoder provides the most accurate bi-partitioning result compared to others.
However, in spite of the improper patch clustering, each layer can build plausible informed mask (and often proper) based on our similarity-score-based masking strategy.
With a simple image, \emph{e.g.}, (a), all layers are able to properly bi-partition the image leading to fully mask out the main object.
With more complex images like (b), (c), (e) and (f), whether 1) the bi-partitioned cluster contains a mixture of foreground and background or 2) only some patches of the foreground are discriminated, our method stably constructs the proper informed mask, aligning with the result in \cref{fig:appendix_mask}.
In example (d), when the layer 7 is used, we observe that  object-centric masks are successfully generated since the pattern of \emph{lizard} is similar to that of \emph{plants}, despite a failure in bi-partitioning where the discriminated foreground captures only the \emph{plants}, missing the \emph{liazard}.
Also, although the decoder hardly captures the entire shape of the foreground, it precisely discriminates the salient patches belonging to the main objects, as expected to generate more accurate informed mask than the early or intermediate encoder layers.
We also note that using the last layer of the encoder yields similar performance to our default setting (\emph{i.e.}, using the second last layer of the encoder) which could be preferred for its simplicity.

As shown in the second group in the \cref{tab:ablation_full}, it is essential to provide hint tokens for successful training.
As displayed in \cref{tab:loss_rebuttal}, the training process results in too high loss without the hint tokens, while it is appropriately alleviated with them.
This is because it is fundamentally impossible for the model to reconstruct the whole foreground without any visible tokens belong to it.
In the aspect of patch clustering, MAE would lose an opportunity to construct high-level latent variables \emph{i.e.}, shared information, for the clusters specific to the foreground when trained without hint tokens.

\input{sec/table_latex/recon_loss_appendix}

In addition to the ablation studies in \cref{tab:ablation}, we also consider 1) the target cluster to be masked out and 2) masking ratio in the third and fourth group in \cref{tab:ablation_full}, respectively.
Object-centric informed masking leads to better performance compared to background-centric masking or alternately masking foreground and background along the epochs, supporting our choice of object-centric masking strategy in \cref{sec:approach}.
For the masking ratio, although masking less regions (0.6) yields the same linear probing performance to the default one (0.75), it is recommended to set masking ratio to 0.75 for more efficient training cost.

\input{sec/figure_latex/appendix_layer_comparison}

\section{Comparison with Various MAEs}
\label{sec:exp:baseline}

We compare the performance across various MAEs in \cref{tab:baseline_rebuttal}, grouping the models based on the incorporation of external training costs.
For models that we were able to reproduce the results
(AMT~\cite{liu2023amt}, HPM~\cite{wang2023hpm}), we report the reproduced results. 
For SemMAE~\cite{li2022semmae}, we refer the performance as reported in the paper.
For CL-MAE~\cite{madan2024clmae}, we report only the training time, as its reproduction is difficult due to the high training cost, and the results reported in the respective paper are not directly comparable due to different experimental settings.

\input{sec/table_latex/baselines_appendix}

\section{Extended Training}
\label{sec:exp:scalability}
We conduct extended pre-training sessions and report linear probing performance on ImageNet-1K along the training epochs in \cref{tab:epoch_scalability}. Our method consistently brings sustained performance gain after considerable length of training, \emph{i.e.}, for 1600 epochs.
\input{sec/table_latex/scalability}

\section{Computing Resources}
\label{appendix:cost}
We conduct experiments on 8 NVidia A6000 GPUs (48GB) and it takes \textasciitilde2.5 days on pre-training for 400 epochs. For 1600 epochs of pre-training, it takes about 10 days.

%% file: sec/figure_latex/hierarchical_appendix.tex
\begin{figure}[h]
    \centering
    \includegraphics[width=0.7\linewidth]{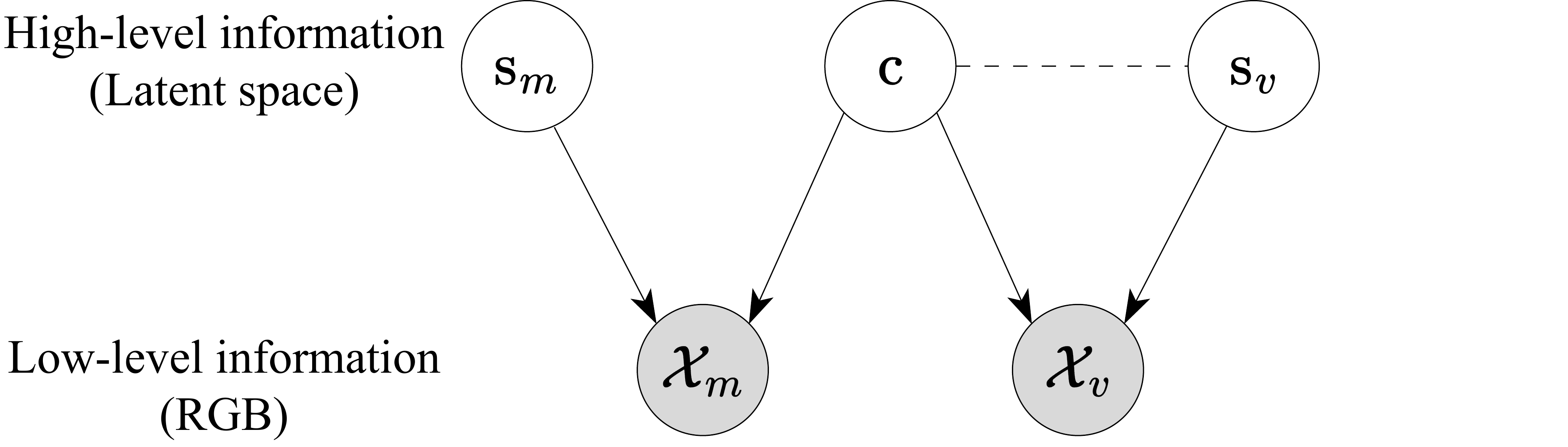}
    \caption{\textbf{Hierarchical latent variable model framework~\cite{kong2023hierarchical}.} 
    Assuming high-level shared information $\mathbf{c}$ exists among the whole tokens, MAE encoder learns to estimate $\hat{\mathbf{c}}$ from $\mathcal{X}_v$ to reconstruct raw pixels of $\mathcal{X}_m$. Here, shared information is equivalent to statistical dependency inside $\mathcal{X}$. $\mathbf{s}_m$ and $\mathbf{s}_v$ stand for information specific to $\mathcal{X}_m$ and $\mathcal{X}_v$, respectively. Dotted line indicates potential dependency.}
    \label{fig:hierarchical_appendix}
\end{figure}

%% file: sec/figure_latex/appendix_mask.tex
\begin{figure*}
    \centering
    \includegraphics[width=\linewidth]{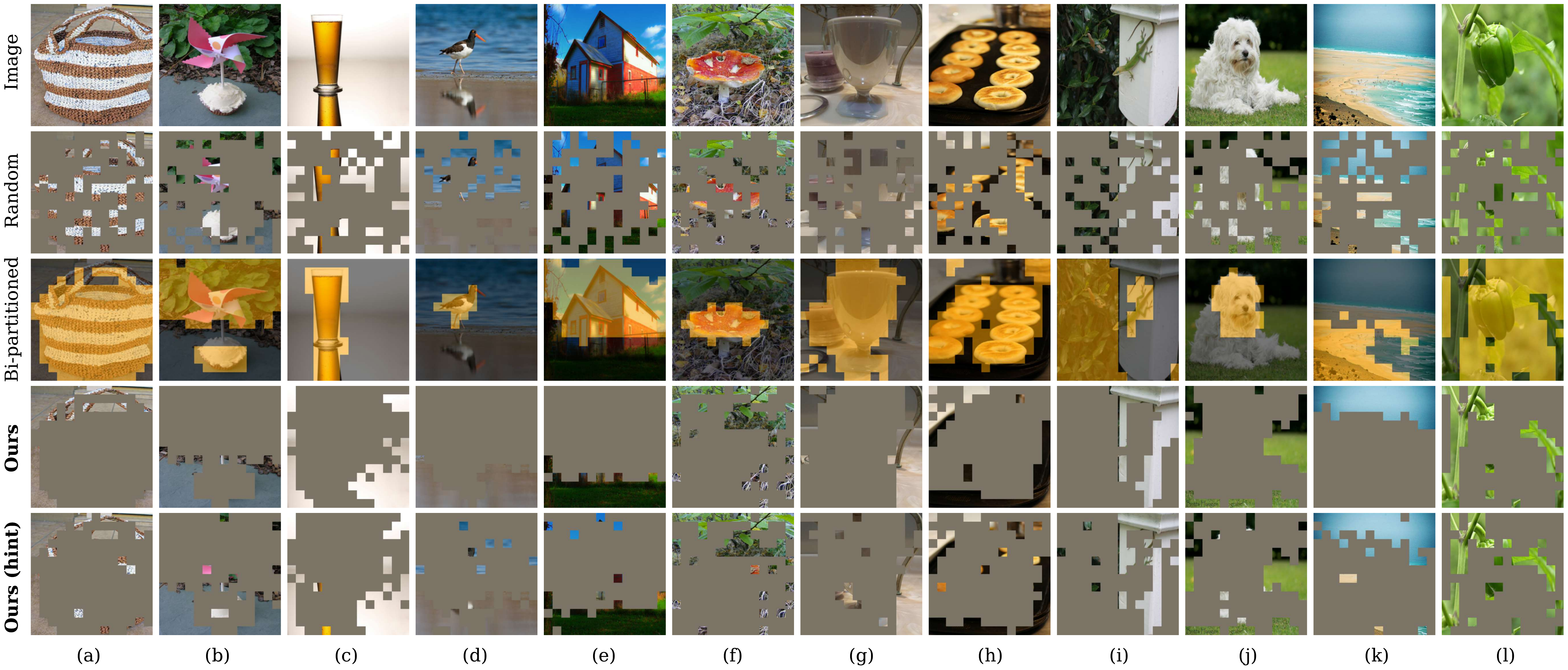}
    \caption{\textbf{Qualitative examples of informed masking on ImageNet training set.} Based on our method, informed masks are generated after 51 epochs of pre-training with a hint ratio of 0.05. Results clearly show that MAE in early training steps provides appropriate bi-partitioning information and successfully creates informed mask without using external models or additional information. We also note that, our similarity-score-based masking strategy yields robust informed mask even in the case when the bi-partitioning is imperfect.}
    \label{fig:appendix_mask}
\end{figure*}

%% file: sec/figure_latex/mae_tsne.tex
\begin{figure}[h]
    \centering
    \includegraphics[width=1.\linewidth]{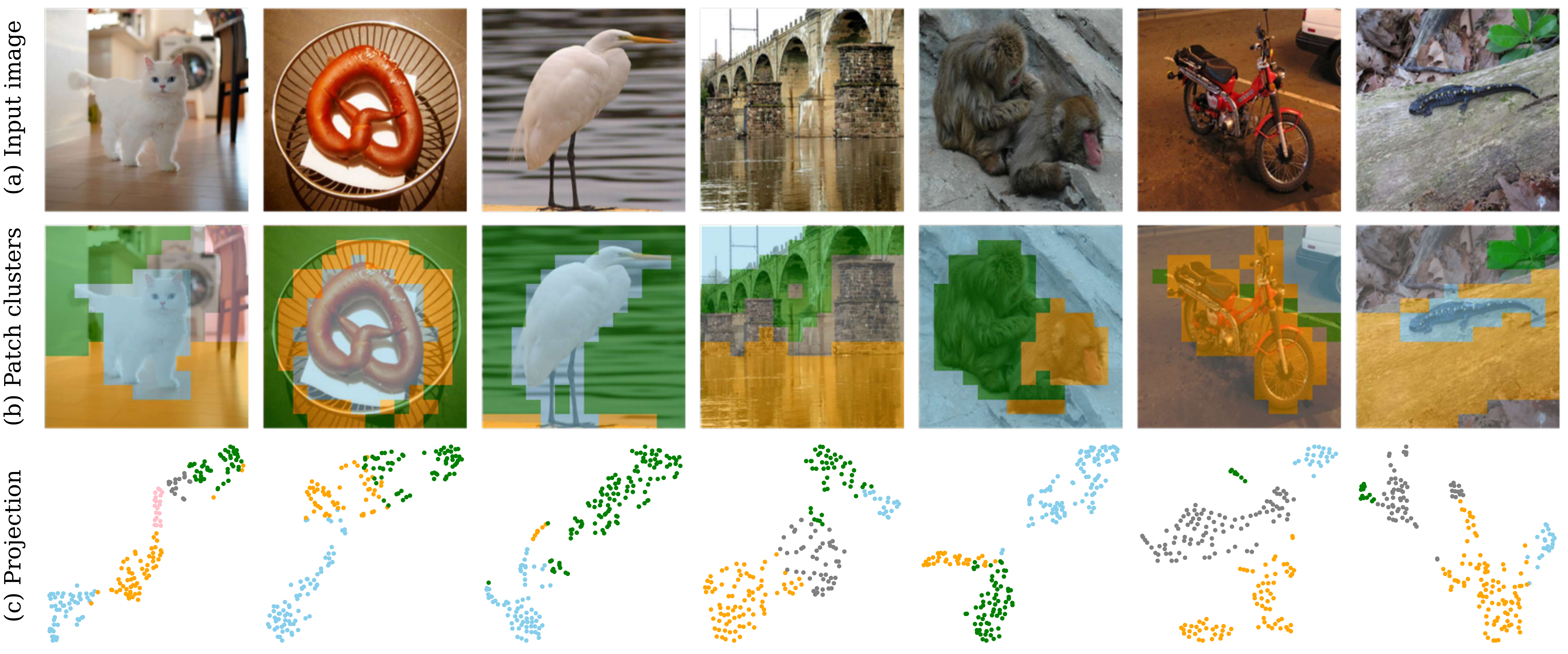}
    \vspace{-0.0cm}
    \caption{\textbf{Illustrations of patch clusters learned by MAE.} (a) Input images. (b) Similarity-based patch clusters. (c) t-sne plots of the patch embeddings.}
    \label{fig:mae_tsne}
\end{figure}



%% file: sec/table_latex/feat_var_appendix.tex

\newcolumntype{C}[1]{>{\centering\arraybackslash}p{#1}}
\begin{wraptable}{r}{.41\linewidth}
\vspace{-0.43cm}
\centering
    \captionsetup{width=.95\linewidth}
    \caption{Feature variance ($\sigma_F$) and similarity variance ($\sigma_S$).}
    \vspace{-0.1cm}
    \footnotesize
    \begin{tabular}{l|C{1.25cm}|C{1.25cm}} 
        \toprule
        Feature & $\mathbb{E}[\sigma_F]$ & {$\mathbb{E}[\sigma_S]$} \\ 
        \midrule
        MAE~\cite{he2022masked} & 0.082 & 0.075 \\
        Ours & 0.096 & 0.079 \\
        \bottomrule
    \end{tabular}
    \label{tab:variance_rebuttal}
\vspace{-.4cm}
\end{wraptable}

%% file: sec/figure_latex/appendix_kld.tex
\begin{figure}[h]
    \includegraphics[width=\linewidth]{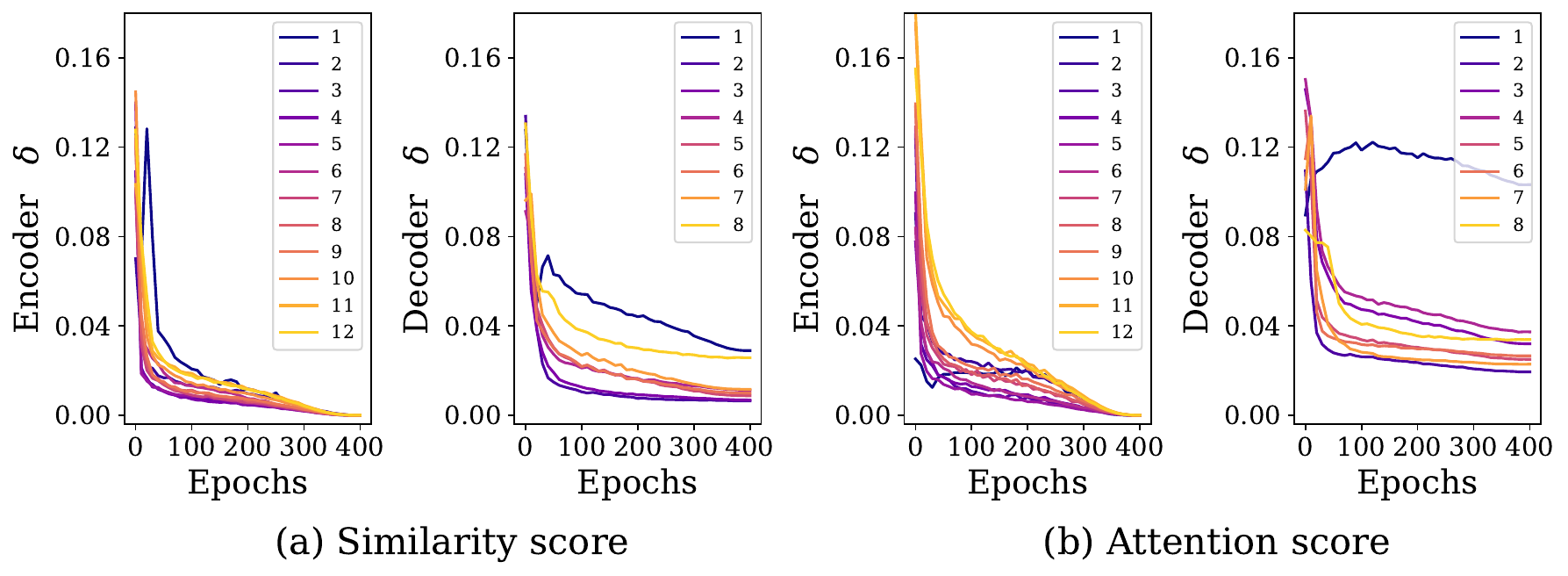}
    \caption{\textbf{KL divergence of the token relations between the final and intermediate epochs.} 
    Layer numbers are displayed in the legend. All the layers but the first one in decoder show drastic decrement of (a) similarity score and (b) attention score at early epochs. The convergence speed and the final converged values vary in layers.}
    \label{fig:appendix_kld}
\end{figure}

%% file: sec/figure_latex/appendix_bipartitionings.tex
\begin{figure}
    \centering
    \includegraphics[width=\linewidth]{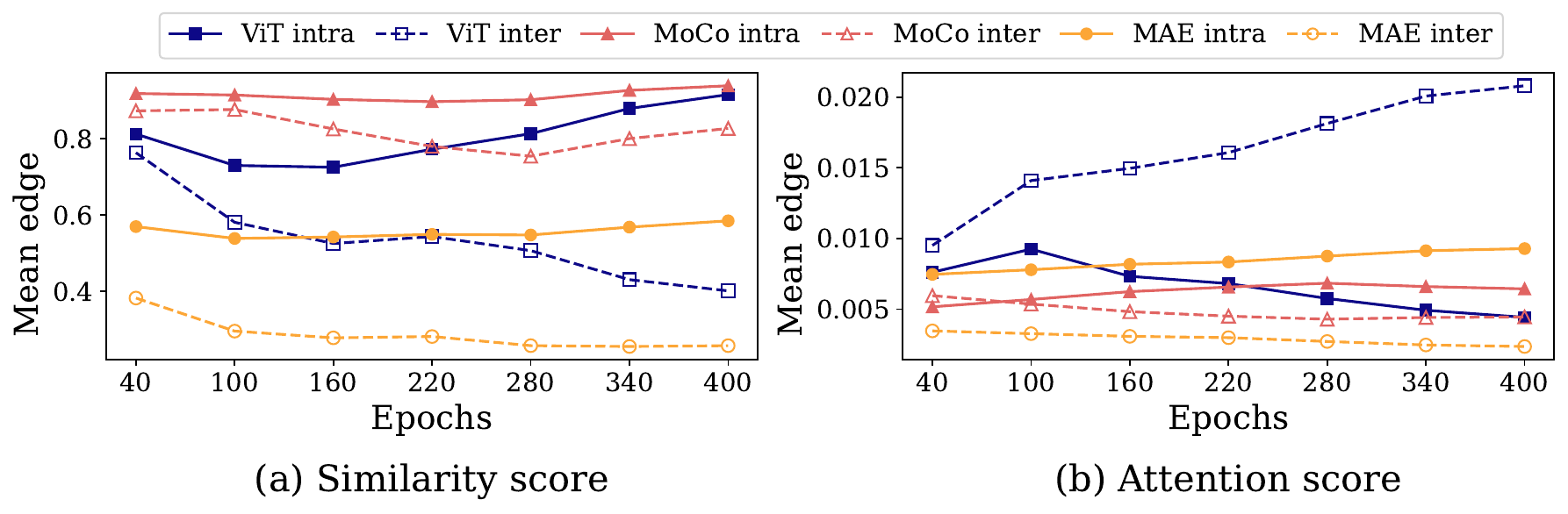}
    \caption{\textbf{Bi-partitioning performance of various models.} 
    MAE, MoCo and ViT show different trends of bi-partitioning performance in both of (a) similarity score and (b) attention score.}
    \label{fig:appendix:bipartitioning}
\end{figure}

%% file: sec/figure_latex/appendix_kld_moco_vit.tex
\begin{figure}[h]
    \includegraphics[width=\linewidth]{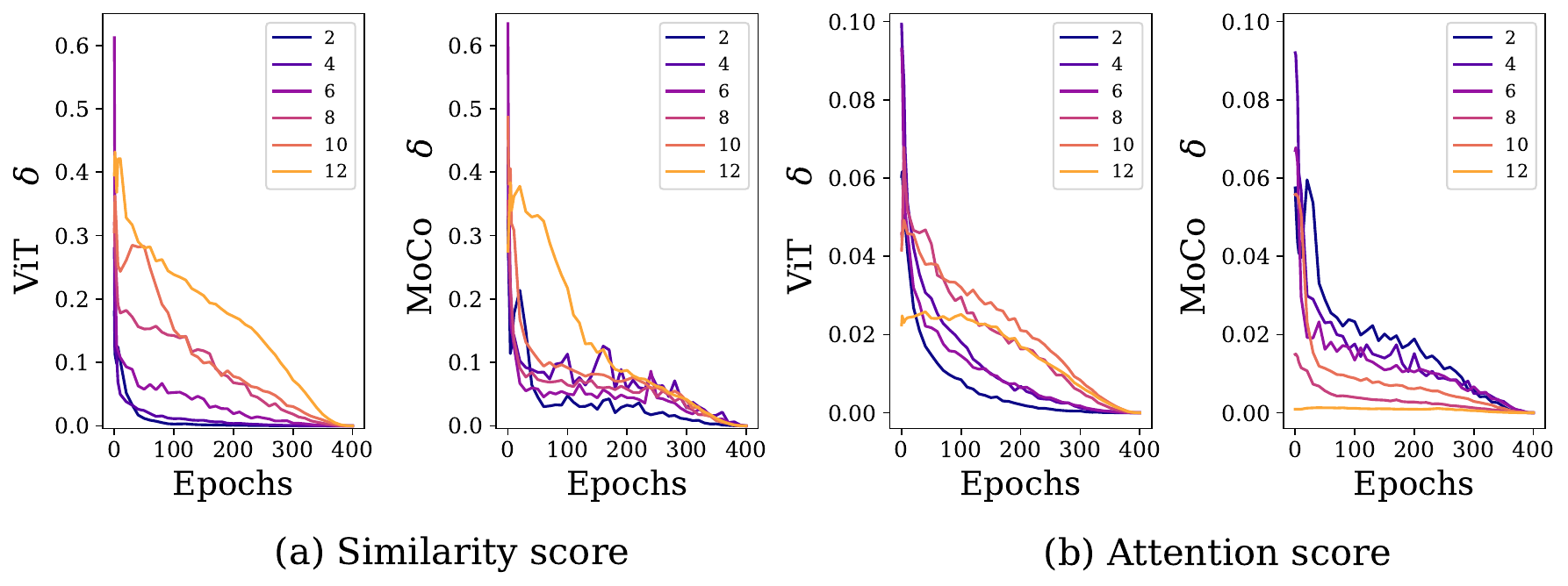}
    \caption{\textbf{KL divergence of token relations of various models.} 
    MoCo and ViT show weaker convergence of token relations in both of (a) similarity score and (b) attention score.}
    \label{fig:appendix:kld_moco_vit}
\end{figure}

%% file: sec/figure_latex/appendix_qualitative.tex
\begin{figure*}
    \centering
    \includegraphics[width=\linewidth]{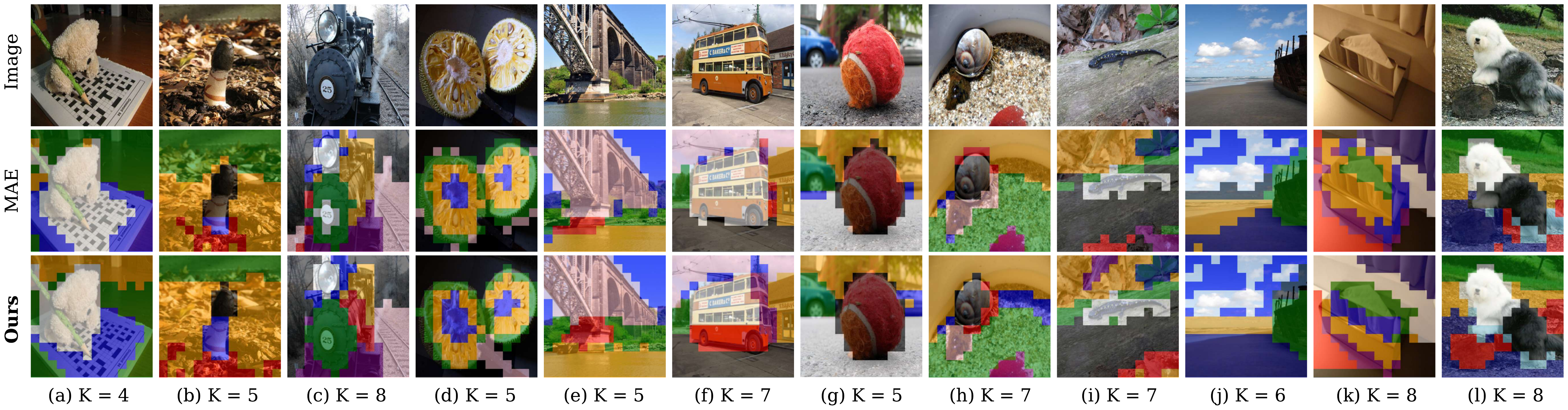}
    \caption{\textbf{Qualitative comparison on ImageNet validation set.} 
    Patches are discriminated in more fine-grained manner with our method. More diverse and finer patch clusters constructed in foreground verify our hypothesis that intensive masking on specific cluster leads to establish more diverse high-level latent variables.}
    \label{fig:appendix_qual}
\end{figure*}

%% file: sec/table_latex/ablation_full_appendix.tex
\newcolumntype{Y}{>{\centering\arraybackslash}X}
\definecolor{lightgray}{gray}{0.9}
\begin{table}[h]
    \centering
    \captionsetup{width=.9\linewidth}
    \caption{\textbf{Ablation studies on various factors.} The default is highlighted in gray.}
    \vspace{.1cm}
    \footnotesize
    \renewcommand{\tabcolsep}{5pt}
    \resizebox{0.9\linewidth}{!}{
    \begin{tabular}{C{1.4cm}C{2.4cm}C{2.4cm}C{2.4cm}C{2.4cm}}
        \toprule
        Layer & Target cluster & Hint strategy & Masking ratio & Linear probing \\
        \midrule
        Enc 3 & Object & Random & 0.75 & 62.3 \\
        Enc 7 & Object & Random & 0.75 & 62.4 \\
        Dec 8 & Object & Random & 0.75 & 62.7 \\
        \midrule
        Enc 11 & Object & $S_i$-based & 0.75 & 62.5 \\
        Enc 11 & Object & No hint & 0.75 & 52.3\\
        \midrule
        Enc 11 & Background & Random & 0.75 & 61.1\\
        Enc 11 & Alternate & Random & 0.75 & 61.6\\
        \midrule
        Enc 11 & Object & Random &  0.6 & \underline{62.9}\\
        Enc 11 & Object & Random &  0.9 & 61.4\\
        \rowcolor{lightgray}
        \midrule
        Enc 11 & Object & Random & 0.75 & \textbf{62.9}\\
        \bottomrule
    \end{tabular}}
    \label{tab:ablation_full}
\end{table}

%% file: sec/table_latex/recon_loss_appendix.tex

\begin{table}[h]
    \centering
    \captionsetup{width=.7\linewidth}
    \caption{\textbf{Reconstruction loss (MSE)} with 400 pre-training epochs according to each training method.}
    \vspace{0.1cm}
    \renewcommand{\tabcolsep}{8pt}
    \footnotesize
    \begin{tabular}{l|ccc} 
        \toprule
        Model & MAE~\cite{he2022masked} & Ours (no hint) & Ours (with hint) \\
        \midrule
        Loss & 0.41 & 0.64 & 0.56 \\ 
        \bottomrule
    \end{tabular}
    \label{tab:loss_rebuttal}
\end{table}

%% file: sec/figure_latex/appendix_layer_comparison.tex
\begin{figure*}[t]
    \centering
    \includegraphics[width=\linewidth]{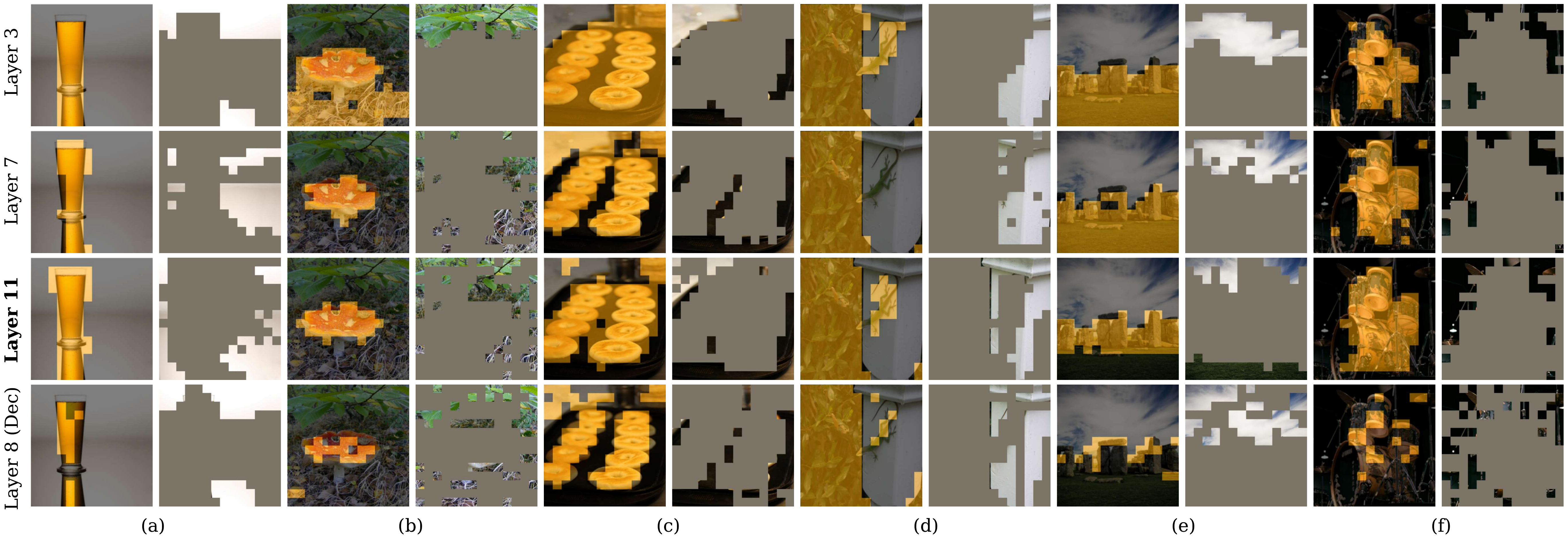}
    \caption{\textbf{Comparison of the Quality of the informed masks generated from different layers.} Each example is denoted by the index of the original image in \cref{fig:appendix_mask}.
    Although early layers of the encoder and the last layer of the decoder yield inappropriate bi-partitioning result, our similarity-score-based masking strategy robustly alleviates this issue, leading to minor difference in performance in the layer selection for generating informed mask.}
    \label{fig:appendix_layer}
\end{figure*}

%% file: sec/table_latex/baselines_appendix.tex
\begin{table}[h]
    \centering
    \caption{\textbf{Comparison with additional MAEs in terms of linear probing performance.} MAEs that utilize external resources or additional parameterized modules are highlighted in gray, indicating that they are not included as baselines for a fair comparison. The training time is reported as a relative value to the MAE, training for 400 epochs. The results show that our method matches the performance of other MAEs with the same or even less training cost. Our method requires only about one more step to generate masks, which empirically increases the pre-training time about 0.25\% for training 400 epochs, \emph{i.e.,} 1.0025$\times$ training time compared to the vanilla MAE.}
    \vspace{0.2cm}
    \renewcommand{\tabcolsep}{5pt}
    \resizebox{.8\linewidth}{!}{
    \begin{tabular}{lcccc} 
        \toprule
        Method & \# of params. & Pre-train epochs & Linear probing & Training time\\
        \midrule
        \multicolumn{4}{l}{\emph{Baselines using external pre-trained model}} \\
        \textcolor{gray!60}{SemMAE~\cite{li2022semmae}} & \textcolor{gray!60}{112M} & \textcolor{gray!60}{800} & \textcolor{gray!60}{68.7} & \textcolor{gray!60}{6.3$\times$} \\
        MAE~\cite{he2022masked} & 112M & 800 & 63.8 & 2$\times$\\
        MAE~\cite{he2022masked} & 112M & 1600 & 68.0 & 4$\times$\\
        \rowcolor{gray!20}
        Ours & 112M & 800 & \textbf{65.9} & 2$\times$\\
        \rowcolor{gray!20}
        Ours & 112M & 1600 & \textbf{68.7} & 4$\times$\\
        \midrule
        \multicolumn{4}{l}{\emph{Baselines using additional module}} \\
        \textcolor{gray!60}{HPM~\cite{wang2023hpm}} & \textcolor{gray!60}{138M} & \textcolor{gray!60}{400} & \textcolor{gray!60}{63.2} & \textcolor{gray!60}{1.5$\times$}\\
        \textcolor{gray!60}{CL-MAE~\cite{madan2024clmae}} & \textcolor{gray!60}{148M} & \textcolor{gray!60}{400} & \textcolor{gray!60}{-} & \textcolor{gray!60}{6$\times$}\\
        \midrule
        \multicolumn{4}{l}{\emph{Baselines \textbf{without} external resource or additional module}} \\
        AMT~\cite{liu2023amt} & 112M & 400 & 61.7 & 1$\times$\\
        MAE & 112M & 400 & 61.4 & 1$\times$\\
        \rowcolor{gray!20}
        Ours & 112M & 400 & \textbf{62.9} & 1$\times$ \\
        \bottomrule
    \end{tabular}}
    \label{tab:baseline_rebuttal}
\end{table}

%% file: sec/table_latex/scalability.tex
\begin{table}[h]
    \centering
    \caption{Linear probing with ImageNet-1K}
    \vspace{0.2cm}
    \small
    \renewcommand{\tabcolsep}{5pt}
    \resizebox{0.65\linewidth}{!}{
    \begin{tabular}{l|C{1.2cm}C{1.2cm}C{1.2cm}C{1.2cm}} 
        \toprule
        Pre-training epochs~~ & 200 & 400 & 800 & 1600 \\
        \midrule
        MAE~\cite{he2022masked} & 53.9 & 61.4 & 63.8 & 68.0\\ 
        Ours & \textbf{54.4} & \textbf{62.9} & \textbf{65.9} & \textbf{68.7} \\ 
        \bottomrule
    \end{tabular}}
    \label{tab:epoch_scalability}
\end{table}